  \RequirePackage{fix-cm}
  \documentclass[twocolumn]{svjour3}          
  \smartqed  
\usepackage{epsfig}
\usepackage{xspace}
\usepackage{color}
\usepackage{multirow}
\usepackage{adjustbox}
\usepackage{subfigure}
\usepackage{amsmath}
\usepackage[noend]{algpseudocode}
\usepackage{algorithmicx,algorithm}
\usepackage{paralist} 
\usepackage{enumitem}
\usepackage{graphicx} 
\usepackage{epstopdf}
\usepackage{booktabs} 

\usepackage[pagebackref=true,breaklinks=true,letterpaper=true,colorlinks,citecolor=blue,linkcolor=blue,urlcolor=blue, bookmarks=false]{hyperref}

\long\def\ignorethis#1{}

\definecolor{Gray}{rgb}{0.35,0.35,0.35}
\definecolor{Blue}{rgb}{0,0.2,0.8}
\definecolor{Red}{rgb}{0.8,0.2,0}
\definecolor{Green}{rgb}{0.0,0.5,0.1}
\definecolor{Gray}{rgb}{0.4,0.4,0.4}

\newlength\paramargin
\newlength\figmargin
\newlength\secmargin

\usepackage{array}
\newcolumntype{L}[1]{>{\raggedright\let\newline\\\arraybackslash\hspace{0pt}}m{#1}}
\newcolumntype{C}[1]{>{\centering\let\newline\\\arraybackslash\hspace{0pt}}m{#1}}
\newcolumntype{R}[1]{>{\raggedleft\let\newline\\\arraybackslash\hspace{0pt}}m{#1}}

\newcommand{\figref}[1]{Figure~\ref{fig:#1}}
\newcommand{\tabref}[1]{Table~\ref{tab:#1}}

  \usepackage{graphicx, ulem, multirow, amssymb, tabularx, paralist, array, tabu, color, subfigure, bbding, adjustbox}
  \usepackage{natbib}

  \makeatletter
  \newcommand{\thickhline}{%
      \noalign {\ifnum 0=`}\fi \hrule height 1pt
      \futurelet \reserved@a \@xhline
  }
  \newcolumntype{"}{@{\hskip\tabcolsep\vrule width 1pt\hskip\tabcolsep}}
  \makeatother
\begin{document}
  
  \title{Gated Fusion Network for Degraded Image Super Resolution
  }
  
  
  \author{Xinyi Zhang$^{1^*}$ \and
  Hang Dong$^{2^*}$       \and
          Zhe Hu$^{3}$          \and
          Wei-Sheng Lai$^{4}$   \and
          Fei Wang$^{2}$        \and
          Ming-Hsuan Yang$^{4}$
  }
  
  \institute{
            Xinyi Zhang, Hang Dong \at
              \email{\{jacqueline, dhunter\}@stu.xjtu.edu.cn}
           \and
           Zhe Hu \at
              \email{zhe.hu@hikvision.com}
           \and
           Wei-Sheng Lai, Ming-Hsuan Yang \at
              \email{\{wlai24, mhyang\}@ucmerced.edu}
           \and
           \Envelope~Fei Wang \at
              \email{wfx@mail.xjtu.edu.cn}
           \and
           $^*$~Equally contributed\\
           $^1$~School of Software Engineering, Xi’an Jiaotong University, Xi’an, Shaanxi 710049, China\\
           $^2$~College of Artificial Intelligence, Xi’an Jiaotong University, Xi’an, Shaanxi 710049, China\\
           $^3$~Hikvision Research America, Santa Clara, CA, USA\\ 
           $^4$~Electrical Engineering and Computer Science, University of California, Merced, CA, USA
}
  
  \date{Received: date / Accepted: date}

  \maketitle
  
  \begin{abstract}
  Single image super resolution aims to enhance image quality with respect to spatial content, which is a fundamental task in computer vision.
  In this work, we address the task of single frame super resolution with the presence of image degradation, e.g., blur, haze, or rain streaks.
  Due to the limitations of frame capturing and formation processes, image degradation is inevitable, and the artifacts would be exacerbated by super resolution methods.
  %
To address this problem, we propose a dual-branch convolutional neural network to extract base features and recovered features separately.
  The base features contain local and global information of the input image.
  On the other hand, the recovered features 
  focus on the degraded regions and are used to remove the degradation.
  Those features are then fused through a recursive gate module to obtain sharp features for super resolution.
  %
  By decomposing the feature extraction step into two task-independent streams, the dual-branch model can facilitate the training process by avoiding learning the mixed degradation all-in-one and thus enhance the final high-resolution prediction results.
  We evaluate the proposed method in three degradation scenarios.
  Experiments on these scenarios demonstrate that the proposed method 
  performs more efficiently and 
  favorably against the state-of-the-art approaches on benchmark datasets.
  \keywords{super resolution \and image restoration \and deep learning}
  \end{abstract}

  \section{Introduction}
  \label{sec:intro}
  Single image super resolution (SISR) aims to restore a high-resolution (HR) image from a low-resolution (LR) one, such as those captured from surveillance and mobile cameras.
  The generated HR image can improve the performance of the numerous high-level vision tasks, e.g., object detection \citep{sr+detection}, face recognition \citep{faceSR}, and surveillance applications \citep{surveillance,sr+face}.
  However, image degradation is often inevitable due to the limitations of the imaging processors and complex capturing scenes.
  For example, motion blur, as well as hazy and rainy weather would introduce undesired artifacts in the captured LR images.
  Those artifacts cannot be fully removed by the imaging formation pipeline and would adversely affect the super resolution algorithms and the following high-level tasks.
  The problems of super resolution and image restoration from degradation are often dealt with separately, as each one is known to be ill-posed. %
  However, such a strategy is neither optimal nor efficient due to error accumulation.

  \begin{figure*}[th]
    \centering
    \begin{minipage}[t]{0.47\textwidth}
    \subfigure[Blurry low-resolution input]{
    \includegraphics[width=1\textwidth]{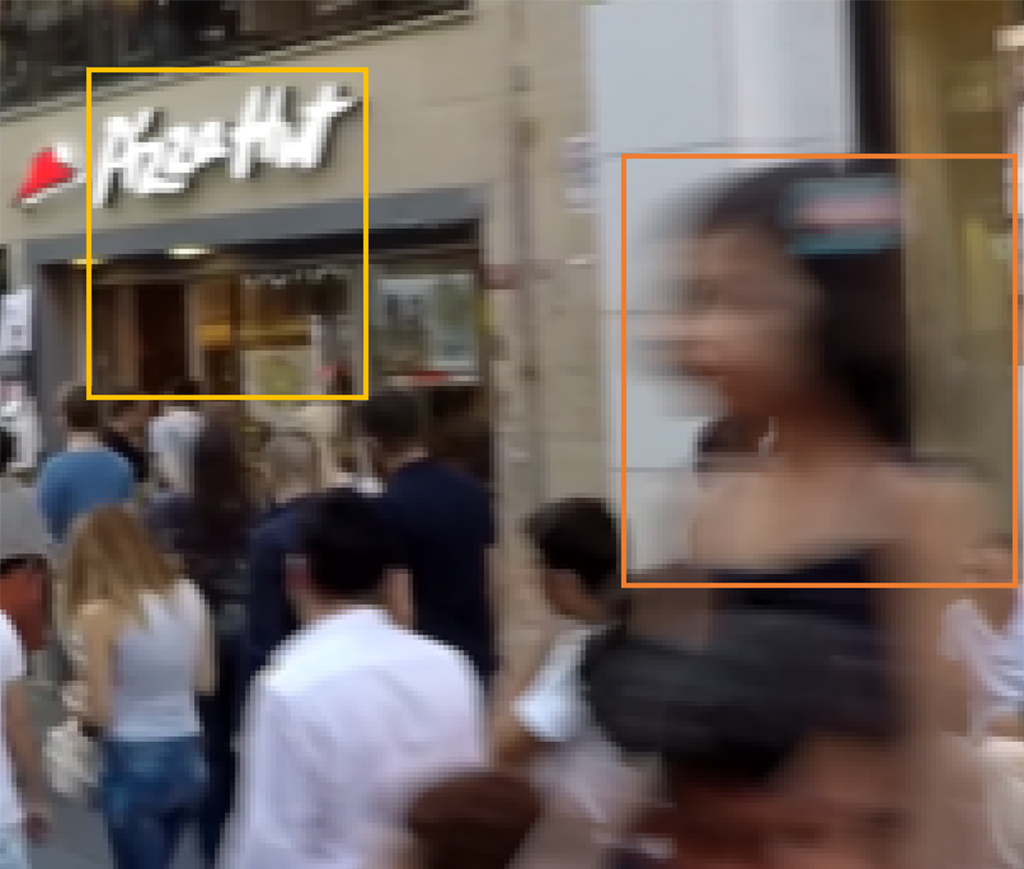} 
    \label{fig:1_all}
    }
    \end{minipage}
    %
    \begin{minipage}[t]{0.513\textwidth}
    \begin{minipage}[t]{0.32\textwidth}
    \subfigure[Input patch]{
    \includegraphics[width=1\textwidth]{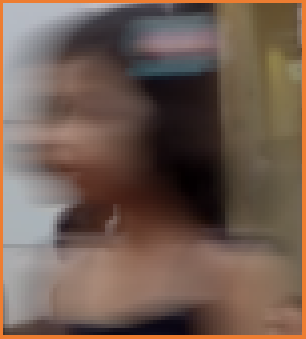} 
    \label{fig:1_sr_input}
    }
    \end{minipage}
    \begin{minipage}[t]{0.32\textwidth}
    \subfigure[\cite{EDSR}]{
    \includegraphics[width=1\textwidth]{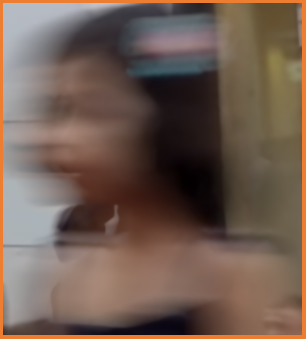} 
    \label{fig:1_sr}
    }
    \end{minipage}
    \begin{minipage}[t]{0.32\textwidth}
    \subfigure[Ours]{
    \includegraphics[width=1\textwidth]{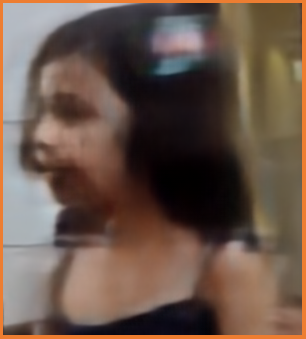}
    \label{fig:1_sr_ours}
    }
    \end{minipage}
    \vfill
    \begin{minipage}[p]{0.32\textwidth}
    \subfigure[Input patch]{
    \includegraphics[width=1\textwidth]{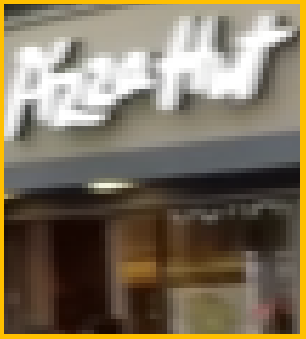} 
    \label{fig:1_db_input}
    }
    \end{minipage}
    \begin{minipage}[p]{0.32\textwidth}
    \subfigure[\cite{deepdeblur}]{
    \includegraphics[width=1\textwidth]{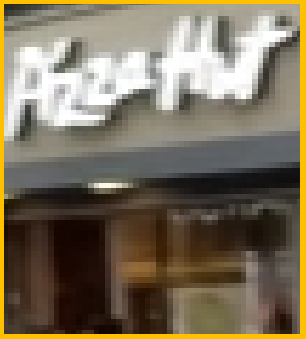} 
    \label{fig:1_db}
    }
    \end{minipage}
    \begin{minipage}[p]{0.32\textwidth}
    \subfigure[Ours]{
    \includegraphics[width=1\textwidth]{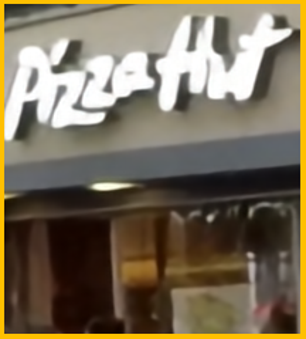} 
    \label{fig:1_db_ours}
    }
    \end{minipage}
    \end{minipage}
    \caption{
    \textbf{Joint image deblurring and super resolution.}
    While the state-of-the-art super resolution algorithm by \cite{EDSR} does not reduce the non-uniform blur in the input image due to the assumption of bicubic downsampling, the top-performing non-uniform deblurring algorithm by \cite{deepdeblur} generates sharp results but with few details.
    In contrast, the proposed model generates a sharp HR image with more details.
    }
    \label{fig:1}
    \end{figure*}
    
  In this work, we address the joint problem of single image super resolution and restoration.
  We evaluate the proposed super resolution architecture on images with three representative image degradations: motion blur, rain streaks, and haze.
  Here, we use super resolution of a blurred image as the example to illustrate this joint task.
  Motion blur is often caused by camera shake, object motion, and scene depth variation.
  \figref{1} shows one blurry LR image, which contains non-uniform blur. 
  As the existing super resolution algorithms \citep{EDSR,srresnet,lapsrn,vdsr} are not designed to handle motion blur explicitly, the resulting HR image is still blurry (see \figref{1_sr_input} and \figref{1_sr}).
  On the other hand, the state-of-the-art non-uniform deblurring methods \citep{non_uniform_deblur1,db&optical,deepdeblur, deblurgan} generate sharp images but cannot restore fine details or enlarge the spatial resolution (see \figref{1_db_input} and \figref{1_db}).

With the advances of deep Convolutional Neural Networks (CNNs), the state-of-the-art image super resolution \citep{EDSR,srresnet,lapsrn} and image restoration \citep{deepdeblur,deblurgan,dehaze12,dehaze10,DID-MDN,RESCAN} methods are developed based on end-to-end networks and achieve promising performance.
To jointly handle the image super resolution and degradation restoration, a straightforward approach is to solve the two sub-proble\-ms sequentially, i.e., performing image restoration followed by super resolution, or vice versa.
However, there are numerous issues within such an approach. 
First, a simple concatenation of two models is prone to error accumulation.
That is, the estimation error of the first model will be propagated and exacerbated in the second model.
Second, the two-step network does not fully exploit the dependence between the two tasks. 
For example, the feature extraction and image reconstruction steps are performed twice and result in computational redundancy.
As both the training and inference processes are memory and time consuming, these approaches cannot be applied to resource-constrained real-time applications, e.g., autonomous driving and video surveillance.
    
Several recent methods \citep{scgan,facesr+attr,icassp18,wenbo} jointly solve the degraded image super resolution problem using end-to-end networks.
However, these methods focus on either domain-specific inputs, e.g., face and text \citep{scgan,facesr+attr} images,
or extending the existing architecture to a particular degradation \citep{wenbo}. 
    
\cite{icassp18} propose a network with two output branches to solve the joint deblurring and super resolution task on natural images. 
Although this method can be extended to super resolve other degraded images by changing the loss functions and training data, it does not perform well when severe degradation exists, e.g., non-uniform blur, heavy rain, or uneven haze.  
    %
    %
In this work, we aim to handle these severe degradations for natural images, which is more challenging.
    \\
    %
    
We use a common image in a dynamic scene
to illustrate the motivation of the proposed method.
The blurry LR input is mixed with degraded (motion blur in this example) regions and relatively sharper regions.
    %
If we extract features from these two regions in one single branch, the training data contains noisy samples and thus makes it difficult to learn an effective model for deblurring.
To address this problem, we propose a \textbf{Gated Fusion Network (GFN)} which consists of two branches: a restoration branch to extract features for recovering the sharp LR image, and a base branch to extract features for fusing.
We adopt a recursive gate module to adaptively fuse the features from two branches for super resolution.
    %
    %
The fused features are then fed into an image reconstruction module to generate the sharp HR image.
Extensive evaluations demonstrate that the proposed model performs favorably against the combination of the state-of-the-art super resolution and image restoration methods as well as the existing joint models in different applications.
    
The contributions of this work are threefold:
    \begin{compactitem}
\item To the best of our knowledge, the proposed method is the first generic deep learning architecture for image super resolution under different degradations.
\item We decouple the joint problem into two sub-tasks for better network regularization.
We propose a dual-branch network to extract the base features and recovered features separately and learn a recursive gate module for adaptive feature fusion.
\item The proposed model entails low computational cost as most operations are performed in the LR space.
Our model performs more efficiently than the combinations of the state-of-the-art super resolution and image restoration methods while achieving significant performance improvement.
\end{compactitem}
    
  \section{Related Work}
\label{sec:realted}
    Both image super resolution and image restoration are fundamental problems in computer vision.
    In this section, we discuss image super resolution and restoration methods closely related to this work.
    
    {\flushleft \bf Image super resolution.}
    Single image super resolution is an ill-posed problem as there are multiple HR images corresponding to the same LR input image.
    Conventional approaches learn the LR-to-HR mappings using sparse dictionaries \citep{A+}, random forest \citep{RFL}, or self-similarity \citep{SelfExSR}.
    In recent years, the CNN-based methods \citep{srcnn,vdsr} have demonstrated significant improvement against conventional super resolution approaches.
    Several techniques have been developed based on recursive learning \citep{drrn,drcn}, pixel shuffling \citep{pixelshuffle,EDSR}, Laplacian pyramid \citep{lapsrn}, back-projection \citep{DBPN}, and channel attention \citep{RCAN}.
    In addition, several approaches use the adversarial loss \citep{srresnet}, perceptual loss \citep{Johnson-ECCV-2016}, and texture loss \citep{enhancenet} to generate super resolution images.
    As most super resolution algorithms assume that the LR images are generated by a simple downsampling kernel, e.g., bicubic kernel, they do not perform well when the input images suffer from other unexpected degradation.
    In contrast, the proposed model is able to super resolve LR images with severe degradation.

    {\flushleft \bf Motion Deblurring.}
    Most existing image deblurring approaches \citep{cho2009fast,xu2013unnatural,pan2016blind,uniform_db1,uniform_db2,uniform_db3} assume that the blur is uniform and spatially invariant across the entire image.
    However, due to depth variation and object motion, real-world images typically contain non-uniform blur.
    Several approaches address the non-uniform deblurring problem by jointly estimating blur kernels with scene depth \citep{kernel_estimate_non-uniform1,hu2014joint} or segmentation \citep{db_related_1}.
    As the kernel estimation step is computationally expensive, recent methods \citep{text_db_cnn,non_uniform_deblur1,deepdeblur,non-uniform_kernel-free} learn deep CNNs to bypass the kernel estimation and efficiently solve the non-uniform deblurring problem.
    \cite{deblurgan} adopt the Wasserstein generative adversarial network (GAN) to generate realistic deblurred images and facilitate the object detection task.
    {\flushleft \bf Image Dehazing.}
    Existing single image dehazing methods often rely on strong image priors or statistical assumptions \citep{dehaze4,dehaze5,dehaze6}.
    \cite{dehaze5} assumes that haze-free images should have higher contrast compared with corresponding hazy images.
    \cite{dehaze6} propose the dark channel prior for haze-free outdoor images and achieve impressive results.
    Recent algorithms \citep{dehaze8,dehaze9,dehaze10} adopt deep CNNs to estimate the transmission map, a major component in the haze model, for reconstructing the haze-free outputs.
    However, inaccurate transmission maps often adversely affect the dehazing results \citep{dehaze9}.
    Therefore, end-to-end architectures have been proposed (e.g., \citep{GFN,dehaze12}) to directly recover the haze-free image without estimating the transmission map.

    {\flushleft \bf Image Deraining.}
    It is challenging to develop restoration algorithms to deal with images captured from outdoor scenes as the contents are complex, dynamic, and with large lighting variations.
    Existing deraining methods can be categorized as video-based \citep{video1,video2,video3} or image-based \citep{SCDL,low-rank,GMM,DID-MDN,ID-CGAN,RESCAN,NLEDN,JORDER,ResGuideNet}. 
    Although video-based algorithms perform better by exploiting the temporal information, 
    the single image deraining problem receives much research attention because of its flexibility and generality.
    Early methods rely on handcrafted low-level features and prior information, e.g., sparse coding and dictionary learning \citep{SCDL}, low-rank representation \citep{low-rank}, and Gaussian mixture models \citep{GMM}.
    However, these schemes are prone to failures of recovering high-frequency details and removing the rain streaks completely.
    Recent approaches show promising improvement based on deep CNNs \citep{DID-MDN,ID-CGAN,NLEDN,ResGuideNet}, recurrent neural network (RNNs) \citep{RESCAN}, and iterative networks \citep{JORDER}.
    \cite{DID-MDN} propose a multi-streaming network for joint rain event detection and deraining.
    Recently, \cite{ID-CGAN} introduce the conditional adversarial loss to recover high-frequency details and a refined loss to suppress the artifacts.
    \cite{RESCAN} utilize a deep convolutional RNN to remove the overlap rain streaks with multiple stages.
 
    \vspace{-2mm}
    {\flushleft \bf Degraded Image Super Resolution.}
    Most super resolution methods in the literature operate on images without significant degradation caused by noise or blur. 
    Some approaches \citep{videoSRDB1,videoSRDB2,videoSRDB3} aim to solve the joint task of super resolution and deblurring by exploiting temporal information from the videos.
    As these methods depend on the optical flow estimation, such schemes cannot be applied to the case of single input images.
    \cite{scgan} train a generative adversarial network to super resolve blurry face and text images.
    As face and text images have distinct structured properties that can be exploited, compact models can be developed to address the joint task of super resolution and deblurring for specific object categories. 
    \cite{icassp18} propose a deep encoder-decoder network (ED-DSRN) for joint image deblurring and super resolution.
    However, the HR images are directly reconstructed from the inputs, which tend to generate unexpected structures in severely degraded regions.
    In this work, we design the network architecture to better extract features in the presence of complex degradations.
    The proposed model has fewer parameters than those of \citep{icassp18} and can generate sharp HR images under different degradations.


  \section{Gated Fusion Network}
  \label{sec:method}
  In this section, we describe the architecture design, training loss functions, and implementation details of the proposed GFN for super resolution on degraded images. 
  
  \begin{figure*}[t]
  \centering
  \includegraphics[width=0.99\linewidth]{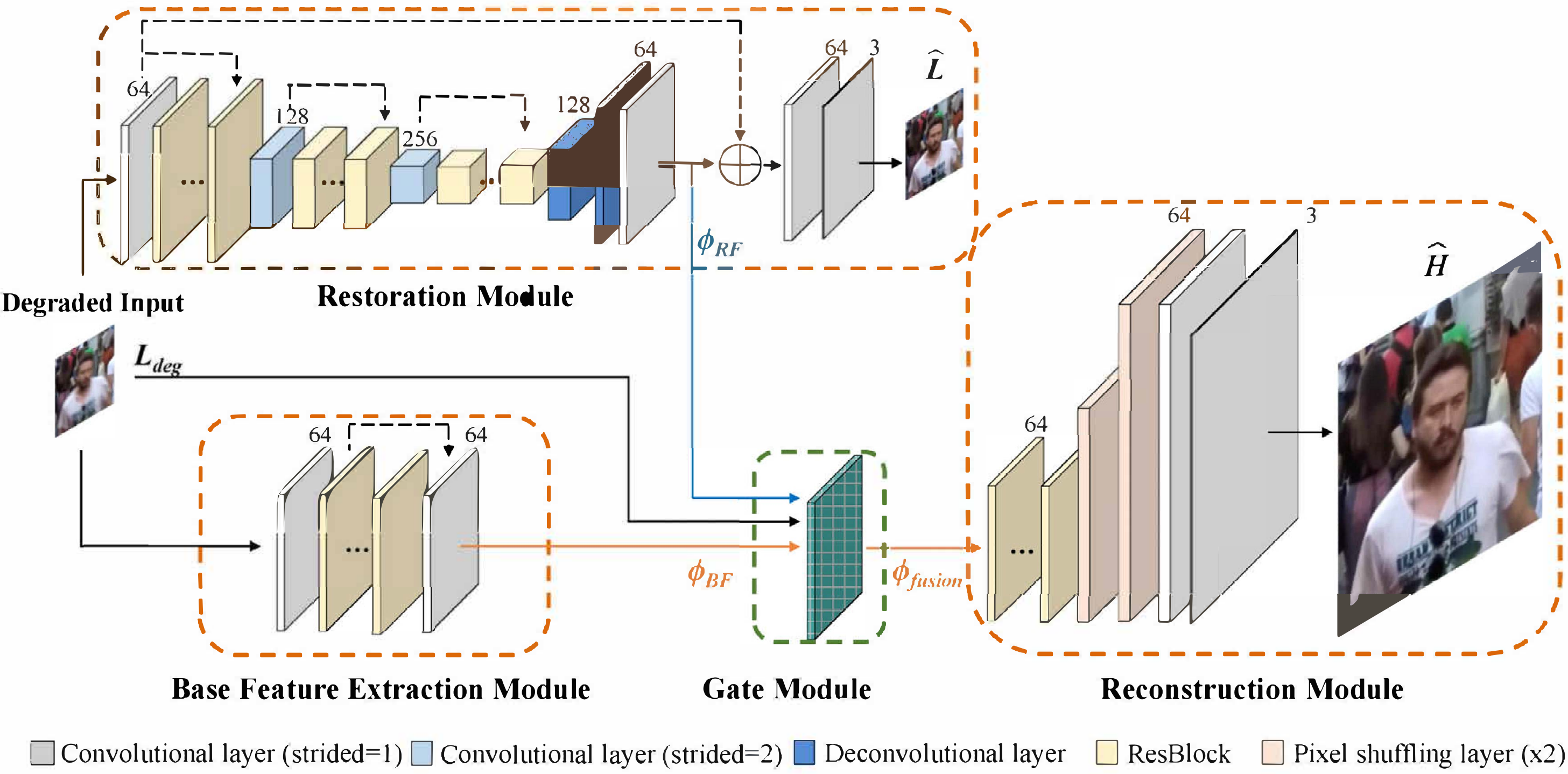}
  \caption{\textbf{Architecture of the proposed GFN model.}
  Our model consists of four major modules:
  restoration module $G_{res}$, base feature extraction module $G_{base}$, gate module $G_{gate}$, and reconstruction module $G_{recon}$.
  The features extracted by $G_{res}$ and $G_{base}$ are fused by $G_{gate}$ and then fed into $G_{recon}$ to reconstruct the HR output image.}
  \label{fig:2}
  \end{figure*}

  \subsection{Network Architecture}
  \label{subsec:architecture}
  Given a degraded LR image $L_{deg}$ as the input, our goal is to recover a sharp HR image $\widehat{H}$.
  In this work, we consider the case of $4$ times super resolution, 
  i.e., the width and height of $\widehat{H}$ are $4$ times larger than those of $L_{deg}$.
  The proposed model has a dual-branch architecture 
  and consists of four major modules:
  (i) a restoration module $G_{res}$ for recovering a sharp LR image $\widehat{L}$, 
  (ii) a base feature extraction module $G_{base}$ to extract visual information from the blurry LR input,
  (iii) a gate module $G_{gate}$ for merging the features from the restoration and base feature extraction modules,
  and (iv) a reconstruction module $G_{recon}$ to reconstruct the final HR output image.
  An overview of the proposed model is illustrated in \figref{2}.

  {\flushleft \bf Restoration Module.}
  The goal of this restoration module is to extract features for recovering a sharp LR image $\widehat{L}$ from the degraded LR input $L_{deg}$.
  We use an asymmetric residual encoder-decoder architecture to enlarge the receptive field.
  The encoder consists of three scales, where each scale has a residual group (six residual blocks as proposed by \cite{EDSR}) and the first two residual groups are followed by a strided convolutional layer to downsample the feature maps by $1/2$ times.
  The decoder has two deconvolutional layers to enlarge the spatial resolution of feature maps.
  Finally, we use two additional convolutional layers to reconstruct a sharp LR image $\widehat{L}$.
  We denote the output features of the decoder by $\phi_{RF}$, which are fed into the gate module for feature fusion.

  {\flushleft \bf Base Feature Extraction Module.}
  We use eight residual blocks \citep{EDSR} to extract base features from the degraded input $L_{deg}$.
  To retain the spatial information, we do not use any pooling or strided convolutional layers.
  We denote the base features by $\phi_{BF}$.
  %
  
\begin{figure}[tb]
  \centering
  \begin{tabular}{ccc}
    \includegraphics[width=0.30\linewidth]{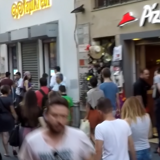} & \hspace{-2mm}
    \includegraphics[width=0.30\linewidth]{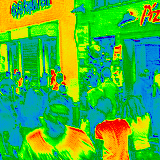} & \hspace{-2mm}
    \includegraphics[width=0.30\linewidth]{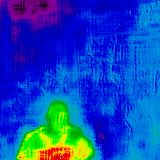} \\
    (a) Input & (b) $\phi_{BF}$ & (c) $\phi_{RF}$       
  \end{tabular}
  \begin{tabular}{ccc} 
    \includegraphics[width=0.30\linewidth]{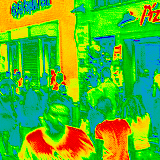} & \hspace{-2mm}
    \includegraphics[width=0.30\linewidth]{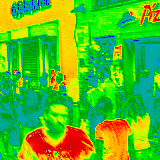} & \hspace{-2mm}
    \includegraphics[width=0.30\linewidth]{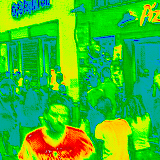} \\
    (c) $\phi{}^{1}_{fusion}$ & (d) $\phi{}^{2}_{fusion}$  & (e) $\phi{}^{3}_{fusion}$ 
  \end{tabular}
  \caption{
  \textbf{Feature responses of the base features $\phi_{BF}$, recovered (deblurring) features $\phi_{RF}$, and fused features from different gate blocks $\phi{}^{1}_{fusion}$, $\phi{}^{2}_{fusion}$, and $\phi{}^{3}_{fusion}$.}
  The base features contain unclear contours around the degraded (blurry) regions, while the recovered (deblurring) features have strong responses on those regions.
  The fused features restore sharp structure and contours information progressively by selectively merging $\phi_{RF}$ into $\phi_{BF}$ in a recursive way.
  We normalize the feature maps for better visualization.}
  \label{fig:Figure3}
  \end{figure}

  {\flushleft \bf Gate Module.}
  In \figref{Figure3}, we show the responses of $\phi_{RF}$, $\phi_{BF}$ and fused features $\phi{}^{n}_{fusion}$ from a blurry LR input.
  While the base features $\phi_{BF}$ contain both sharp and unclear contours (as shown on the wall of \figref{Figure3}(b)), the recovered (deblurring) features $\phi_{RF}$ have high response on the regions with large motion (as shown by the pixels of the moving person in \figref{Figure3}(c)).
  Thus, the responses of $\phi_{RF}$ and $\phi_{BF}$ complement each other, especially on the degraded (blurry) regions.
  To better extract features for super resolution, we adaptively merge the recovered features and base features by learning a gate module, which has been shown effective to discover feature importance for multi-modal fusion \citep{lstm,GFN}.
  We apply a basic gate block and adopt a recursive merging strategy to progressively fuse the features.
  
\begin{figure*}[t]
  \centering
  \includegraphics[width=0.85\linewidth]{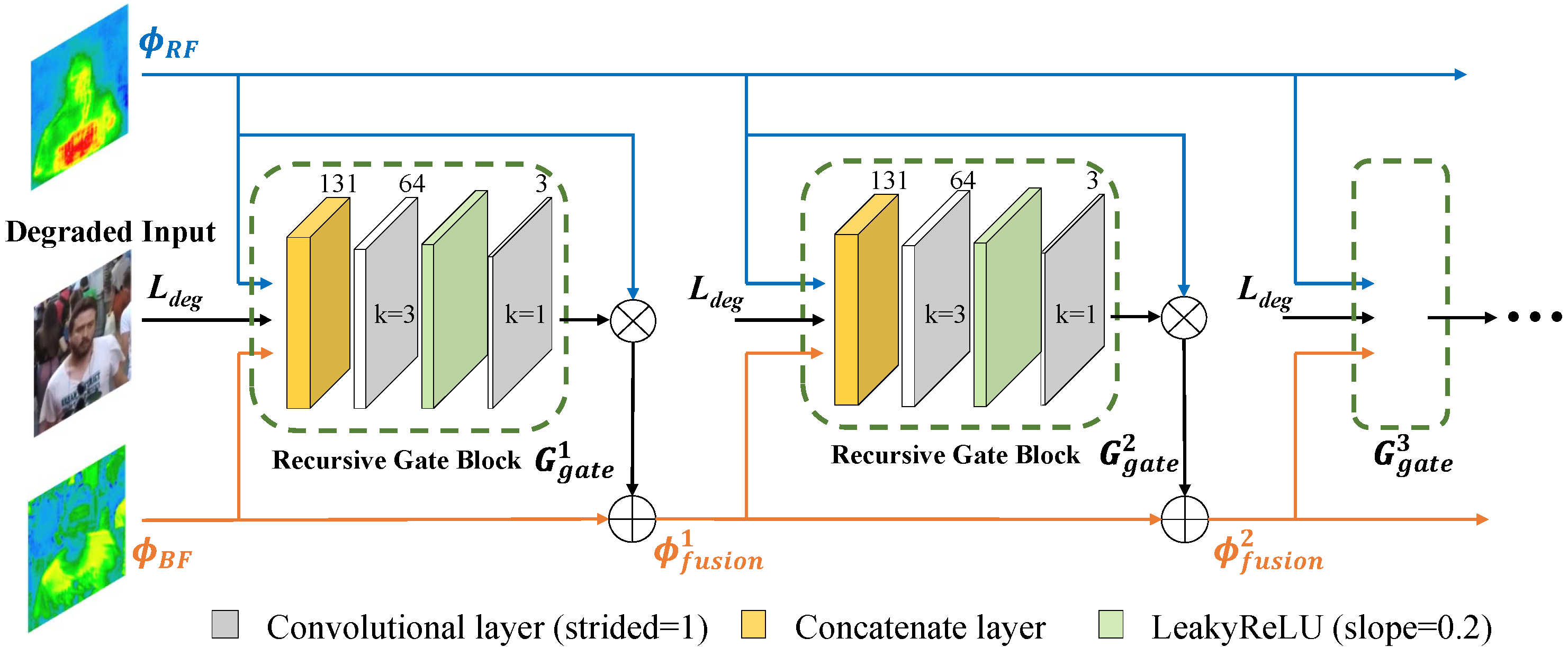}
  \caption{\textbf{Structure of the recursive gate module.}
We use recursive gate blocks to fully exploit the correlation between the features from two independent branches and fuse them progressively.
  Since each block serves as the same purpose of adaptively merging the recovered features $\phi_{RF}$ into the main stream $\phi{}^{n}_{fusion}$  ($\phi_{BF}$ for the first block), the parameters are shared among blocks.
  }
  \label{fig:2-G}
  \end{figure*}

  As shown in \figref{2-G}, each gate block consists of a concatenation layer, two convolutional layers with the filter size of 
  $3 \times 3$ and $1 \times 1$, and a leaky rectified linear unit (LReLU) between the two convolutional layers.
  The first recursive gate block, $G{}^{1}_{gate}$, takes  $\phi_{RF}$, $\phi_{BF}$, and the degraded LR input $L_{deg}$ as input, and generates a pixel-wise weight map.
  The fused features can be formulated as,
  \begin{equation}\label{eqn2}
    \phi{}^{1}_{fusion} = G{}^{1·}_{gate}(\phi_{RF}, L_{deg}, \phi_{BF})  \otimes \phi_{RF} + \phi_{BF}, 
  \end{equation}
  where $\otimes$ denotes the element-wise multiplication.

  We propose a recursive strategy to exploit the dependence of two independent branches for feature fusion.
  We stack $N$ gate blocks, where each block serves as the same purpose of adaptively merging the recovered features $\phi_{RF}$ into the main stream $\phi{}^{n}_{fusion}$ ($\phi_{BF}$ for the first block).
  The parameters are shared among the recursive gate blocks, and the output of the previous block $\phi{}^{k-1}_{fusion}, k=2,\dots,N$ is used as the base features in the next block.
  \figref{Figure3} shows the proposed recursive fusion process.
  Compared to the base features $\phi_{BF}$, the features after the first fusion $\phi{}^{1}_{fusion}$ contain sharp contours of the moving person.
  The fused features after the second and third fusion steps contain clearer and finer information of the person, especially on the chest region, which is useful for HR image reconstruction.

    The DGFN \citep{GFN} method trains a network to predict confidence maps for three hand-crafted enhanced images derived from the input hazy image and then uses a gate module to combine them for generating the sharp image without haze. 
    This method is specifically developed for the single-image dehazing task, which cannot be straightforwardly extended to other restoration tasks due to the usage of the hand-crafted enhanced images.
    In contrast, the proposed method is a generic framework for the joint image restoration and super-resolution problem and does not involve any heuristic process. 
    Our gate module is designed to predict the confidence maps to adaptively fuse the features from two sub-networks, where one extracts features for restoration and another one extracts features from the input image.
    The fused features are then fed into an image reconstruction module to generate the sharp HR output.
  
    To retrieve more contextual information from hand-crafted enhanced images, their gate module is constructed with 3 dilated convolutional blocks and 3 deconvolutional blocks.
    Since our gate module only aims to fuse two extracted features, 
    our gate module only consists of two convolutional layers and a leaky rectified linear unit (LReLU) to maintain simplicity.

  {\flushleft \bf Reconstruction Module.}
  In the final stage, the fused features $\phi{}^{N}_{fusion}$ are fed into eight residual blocks \citep{EDSR} and two pixel-shuffling layers \citep{pixelshuffle} to enlarge the spatial resolution by $4$ times.
  We then use two final convolutional layers to reconstruct an HR output image $\widehat{H}$.
  We note that most of the network operations are performed in the LR feature space.
  Thus, the proposed model entails low computational cost in both training and inference phases.
  
  \subsection{Loss Functions}
  \label{subsec:loss}
  The proposed network generates two output images: a recovered LR image $\widehat{L}$ and a sharp HR image $\widehat{H}$.
  %
  %
  In our training data, each degraded LR image $L_{deg}$ has a corresponding ground truth HR image $H$ and a ground truth LR image $L$, which is bicubic-downsampled from $H$.
  Thus, we train our network by jointly optimizing a super resolution loss and a recovering loss:
  \begin{equation}\label{eqn3}
    \min \mathcal{L}_{SR}(\widehat{H}, H) + \alpha \mathcal{L}_{recover}(\widehat{L}, L),
  \end{equation}
  where $\alpha$ is a weight to balance the two loss terms.

    Without the recovering loss, both the base feature extraction module and restoration module are solely guided by the super resolution loss.
 In this case, there is no guarantee that the dual-branch architecture can learn to extract recovered features.
  We have trained GFN without recovering loss and found that its result in PSNR is worse than the proposed GFN (27.69 vs. 27.91) on the joint deblurring and super resolution problem.
  Therefore, we impose a guidance on the restoration branch using the recovering loss to encourage the branch to extract recovered features for the restoration task.  
  We use the pixel-wise L2 loss function for both $\mathcal{L}_{SR}$ and $\mathcal{L}_{recover}$, and empirically set $\alpha$ to $0.5$.
  
  \subsection{Implementation Details}
  In the proposed network, the filter size is set as $7 \times 7$ in the first and the last convolutional layers, $4 \times 4$ in the deconvolutional layers, $1 \times 1$ in the last convolutional layers of the gate blocks, and 
  $3 \times 3$ in all the other convolutional layers.
  We randomly initialize all the trainable parameters by using the method of \citep{init_HE}.
  We use the leaky rectified linear unit (LReLU) with a negative slope of 0.2 as the activation function.
  As suggested in \citep{EDSR}, we do not use any batch normalization layers in order to retain the range flexibility of features.
To facilitate the training process, we use skip connections in the restoration module and base feature extraction module (refer to the dashed lines in \figref{2}).
  From quantitative evaluations (see \tabref{ablation}), we find that the gate module with 3 recursive gate blocks, i.e., $N=3$, achieves the best performance on all three applications.
  Thus we set $N$ to $3$ as the default parameter of the proposed GFN model.
  We use the ADAM solver \citep{adam} with $\beta_1 = 0.9$ and $\beta_2 = 0.999$ to optimize the network.
  All the training and evaluation processes are conducted on an NVIDIA 1080Ti GPU.
  The source code can be found at \href{https://github.com/BookerDeWitt/GFN-IJCV}{https://github.com/BookerDeWitt/GFN-IJCV}.
  
  
  \section{Experimental Results}
  \label{sec:experiment}
  In this section, we evaluate the proposed GFN model on super resolving blurry, hazy images, and rainy images. 
  We present quantitative and qualitative comparisons with state-of-the-art approaches.
In addition, we carry out ablation studies to analyze several design choices of the proposed model.

  \subsection{Super Resolving Blurry Image}
  \label{subsec:blurrySR}
  {\flushleft \bf Training Dataset and Details.}
  We use the GOPRO \citep{deepdeblur} dataset to generate the training data for the joint super resolution and deblurring problem.
  The GOPRO dataset contains 2103 blurry and sharp HR image pairs for training.
  To augment the training data, we resize each HR image pair with three random scales within the scale of $0.5$ and  $1.0$.
  We then crop the HR images into several patches with a size of $256 \times 256$ and a stride of 128.
  We downsample the blurry HR patch $H_{blur}$ and sharp HR patch $H$ by $4$ times using bicubic downsampling to generate the blurry LR patches $L_{blur}$ and sharp LR patches ${L}$.
  We obtain 107,584 triplets of $\left\{L_{blur}, L, H\right\}$ for training (the blurry HR patches $H_{blur}$ are discarded during training).
  The generated dataset is referred to as LR-GOPRO in the following.

  To facilitate the training process, we use a three-stage training strategy. 
  First, we pre-train the network without the gate module by simply fusing $\phi_{RF}$ and $\phi_{BF}$ via addition.
  Therefore, only the restoration module, base feature extraction module, and reconstruction modules are updated in this stage. 
  We use an initial learning rate of $10^{-4}$ with a decay rate of 0.5 every 6 epochs.
  The network is trained from scratch for 25 epochs.
  We note that the rapidly decaying pre-training without the gate module is important in the early stage as it helps avoid the exploding gradient issues.
  In the second stage, we continue training the models without the gate module for 60 epoch.
  The learning rate is reset to $10^{-4}$ and multiplied by 0.1 for every 30 epochs.
  Finally, we include the gate module and train the entire network for 60 epochs.
  The learning rate is set to $5\times10^{-5}$ and multiplied by 0.1 for every 25 epochs.
  We use a batch size of 16.

  {\flushleft \bf Performance Evaluation.}
  We evaluate the proposed GFN model with the state-of-the-art super resolution methods \citep{srresnet,EDSR,RCAN}, joint image deblurring and super resolution approaches \citep{scgan,icassp18}, and straightforward combinations of super resolution and non-uniform deblurring schemes \citep{deepdeblur,deblurgan,SRN}.
  For fair comparisons, we re-train the SCGAN \citep{scgan}, SRResNet \citep{srresnet}, and ED-DSRN \citep{icassp18} models on the same training dataset discussed above.
  Other super resolution methods are trained on the DIV2K dataset \citep{DIV2K} and deblurring methods are trained on the GOPRO dataset \citep{deepdeblur}.

\begin{figure*}[tb]
  \centering
  \subfigure[PSNR vs. inference time]{\label{fig:psnr_time}
  \begin{minipage}[t]{0.48\linewidth}
    \centering
    \includegraphics[height=5cm]{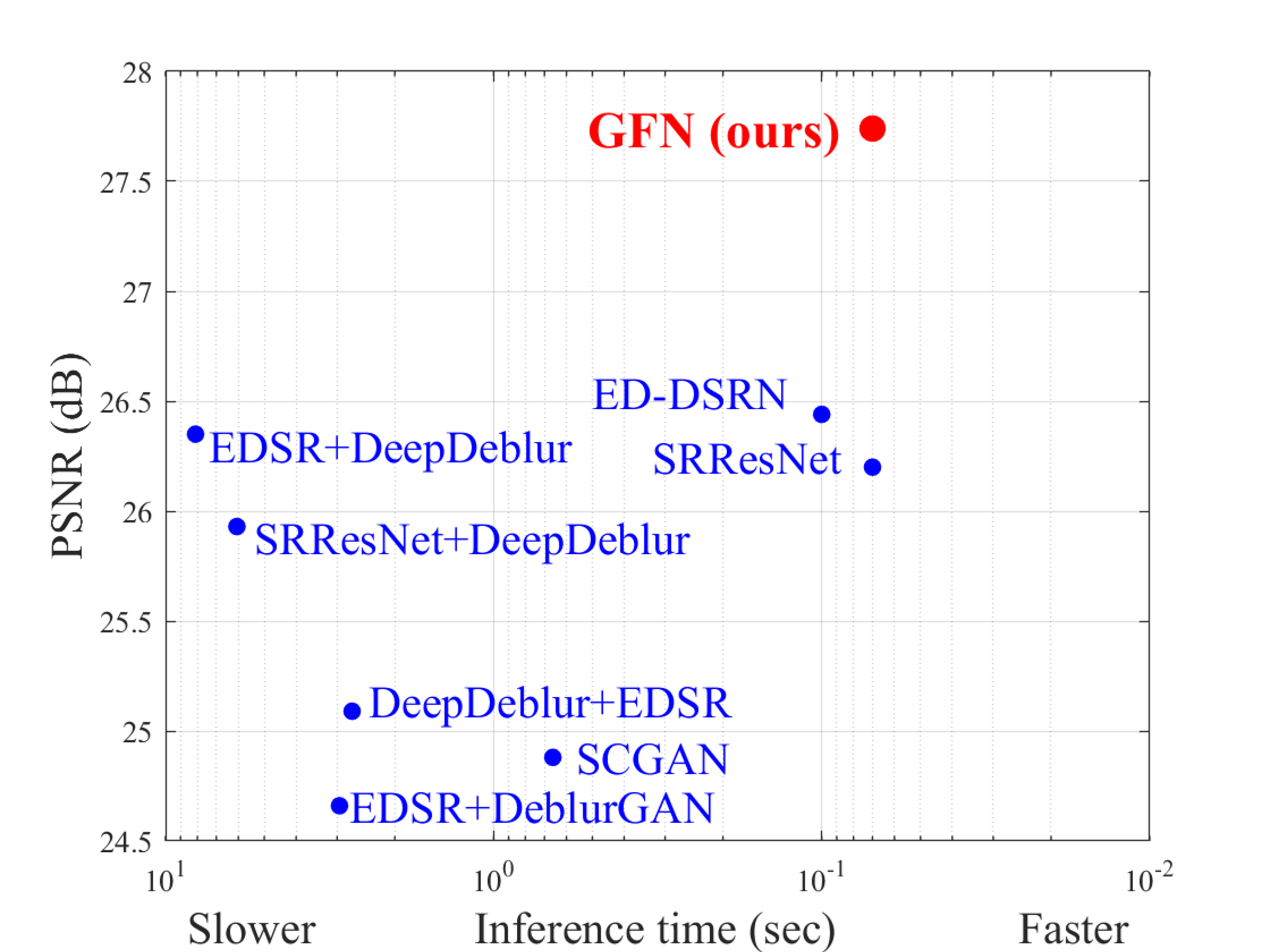}
    \end{minipage}
  }
  \subfigure[PSNR vs. number of parameters]{\label{fig:psnr_params}
  \begin{minipage}[t]{0.48\linewidth}
    \centering
    \includegraphics[height=5cm]{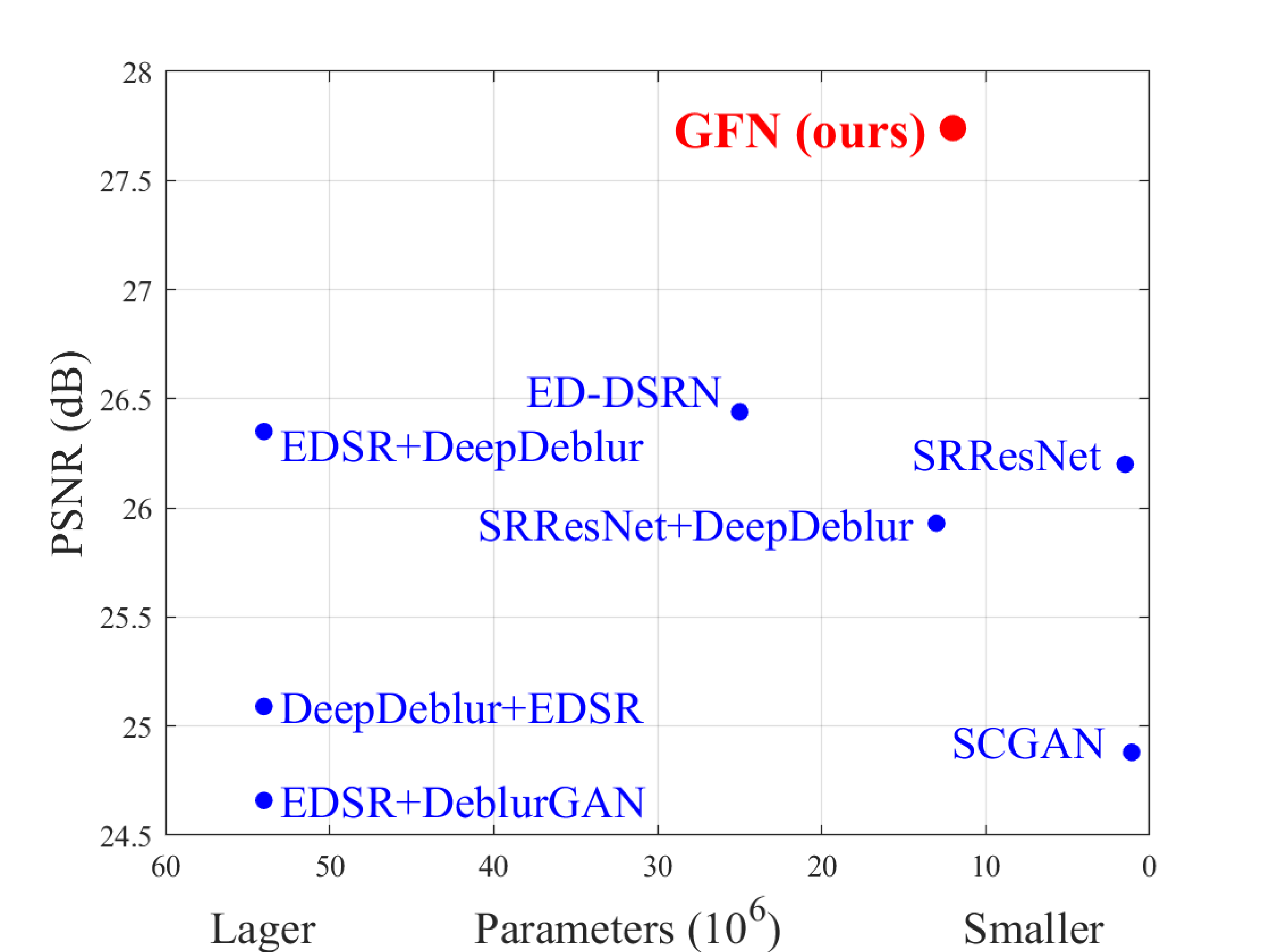}
    \end{minipage}
  }
  \caption{
  \textbf{Performance versus inference time and model parameters.}
  The results are evaluated on the LR-GOPRO dataset.
  }
  \label{fig:psnr_time_params}
  \end{figure*}

  We use bicubic downsampling to generate blurry LR images from the test set of the GOPRO \citep{deepdeblur} and  K\"ohler \citep{kohler} datasets for evaluation.
  \tabref{table_deblur_results} shows the quantitative evaluation in terms of PSNR, SSIM, and average inference time.
  The tradeoff between image quality and efficiency is better visualized in \figref{psnr_time_params}.
  The proposed GFN model performs favorably against existing methods on both datasets and maintains a low computational cost and execution time.
  While the re-trained SCGAN and SRResNet perform better than their pre-trained models, both methods do not handle the complex non-uniform blur well due to their small model capacity.
  It is noted that the SCGAN takes bicubic upsampled images as the inputs, and most operations are performed in the HR feature space.
  In contrast, the ED-DSRN and our GFN take LR images as the inputs, and most operations are performed in the LR feature space.
  Therefore, the SCGAN runs slower than others even with fewer parameters.
  The ED-DSRN method performs well using a large model with more parameters. 
  However, the single-branch architecture of ED-DSRN is less effective than the proposed dual-branch network.
  
  \begin{table*}[t]
  \small
  \centering
  \caption{
  \textbf{Quantitative comparison with the state-of-the-art methods on super resolving the blurry images.}
  The evaluated methods include \textbf{super resolution methods}, SRResNet \citep{srresnet}, EDSR \citep{EDSR}, RCAN \citep{RCAN},
  \textbf{image deblurring methods}, DeepDeblur \citep{deepdeblur}, DeblurGAN \citep{deblurgan}, SRN \citep{SRN},
  and \textbf{joint approaches}, SCGAN \citep{scgan}, ED-DSRN \citep{icassp18}.
  The methods with a $\star$ sign are trained on our LR-GOPRO training set.
  {\color{red}Red texts} indicate the best performance.
  The proposed GFN model performs favorably against existing methods while maintaining a small model size and fast inference speed.
  }
  \label{tab:table_deblur_results}
  \begin{tabular}{rcll}
  \hline
  &\multicolumn{1}{l}{} & \multicolumn{1}{c}{LR-GOPRO $4\times$}   & \multicolumn{1}{c}{LR-K\"ohler $4\times$} 
  \\
  \multirow{-2}{*}{Method}    &\multirow{-2}{*}{\#Params}     & \multicolumn{1}{c}{PSNR~/~SSIM~/~Time~(s)}    & \multicolumn{1}{c}{PSNR~/~SSIM~/~Time~(s)}  
  \\ 
  \hline
  SCGAN        &1.1M      &22.74~/~0.783~/~0.66  &23.19~/~0.763~/~0.45
  \\
  SRResNet  &1.5M    &24.40~/~0.827~/~{\color{red}0.07} &24.81~/~0.781~/~{\color{red}0.05}
  \\
  EDSR          &43M     &24.52~/~0.836~/~2.10    &24.86~/~0.782~/~1.43
  \\
  RCAN          &16M     &24.54~/~0.836~/~1.76   &24.87~/~0.782~/~1.17
  \\
  SCGAN${}^{\star}$	&1.1M &24.88~/~0.836~/~0.66 &24.82~/~0.795~/~0.45
  \\  
  SRResNet${}^{\star}$ &1.5M	&26.20~/~0.818~/~{\color{red}0.07} &25.36~/~0.803~/~{\color{red}0.05}
  \\  
  ED-DSRN${}^{\star}$	&25M &26.44~/~0.873~/~0.10 &25.17~/~0.799~/~0.08
  \\
  \hline
  DeepDeblur + SRResNet &13M &24.99~/~0.827~/~0.66 &25.12~/~0.800~/~0.55
  \\
  SRResNet + DeepDeblur &13M  &25.93~/~0.850~/~6.06 &25.15~/~0.792~/~4.18
  \\
  DeblurGAN + SRResNet  &13M &21.71~/~0.686~/~0.14 &21.10~/~0.628~/~0.12
  \\
  SRResNet + DeblurGAN  &13M &24.44~/~0.807~/~0.91  &24.92~/~0.778~/~0.54
  \\ 
  DeblurGAN + EDSR      &54M &21.53~/~0.682~/~2.18 &20.74~/~0.625~/~1.57
  \\
  EDSR + DeblurGAN  &54M &24.66~/~0.827~/~2.95 &25.00~/~0.784~/~1.92
  \\ 
  DeepDeblur + EDSR     &54M  &25.09~/~0.834~/~2.70 &25.16~/~0.801~/~2.04
  \\
  EDSR + DeepDeblur  &54M &26.35~/~0.869~/~8.10    &25.24~/~0.795~/~5.81
  \\
  DeepDeblur + RCAN &37M &25.10~/~0.833~/~3.91 &25.16~/~0.801~/~3.52
  \\
  RCAN + DeepDeblur &37M &26.34~/~0.870~/~5.39 &25.24~/~0.794~/~4.67
  \\
  SRN + RCAN &17M &25.62~/~0.867~/~3.10 &25.18~/~0.798~/~1.66
  \\
  RCAN + SRN &17M &26.00~/~0.874~/~5.76 &25.20~/~0.799~/~4.49
  \\ 
  \hline
  GFN${}^{\star}$~(ours) &12M & {\color{red} 27.91}~/~{\color{red} 0.902}~/~{\color{red}0.07}    & {\color{red} 25.79}~/~{\color{red} 0.818}~/~{\color{red}0.05}
  \\
  \hline
  \end{tabular}
  \end{table*}

  The straightforward approaches by combining super resolution and deblurring methods are generally less effective due to the error accumulation.
  We note that the approaches first using super resolution (i.e., performing super resolution followed by image deblurring) typically perform better than the alternatives (i.e., performing image deblurring followed by super resolution).
  However, the strategy by first performing super resolution entails heavy computational cost as the time-consuming image deblurring step is performed in the HR image space.
  Compared with the best-performing combination of EDSR and DeepDeblur methods, the proposed GFN model executes $116$ times faster and uses $78\%$ fewer model parameters.

\begin{figure*}[tb]
  \small
  \centering
  \begin{tabular}{cccc}
    \hspace{-3mm}
     \includegraphics[width=0.245\linewidth]{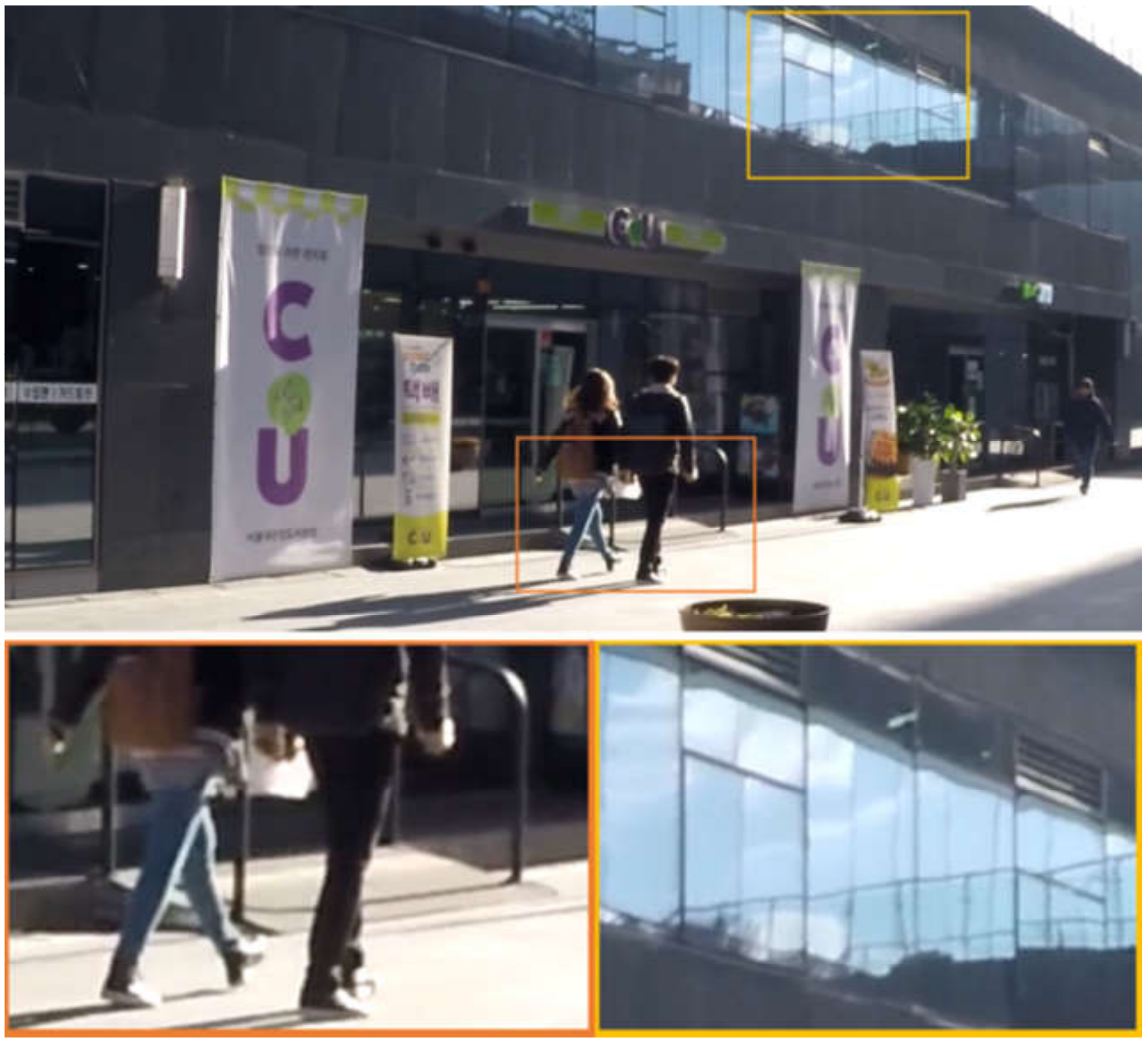} & \hspace{-4mm}
      \includegraphics[width=0.245\linewidth]{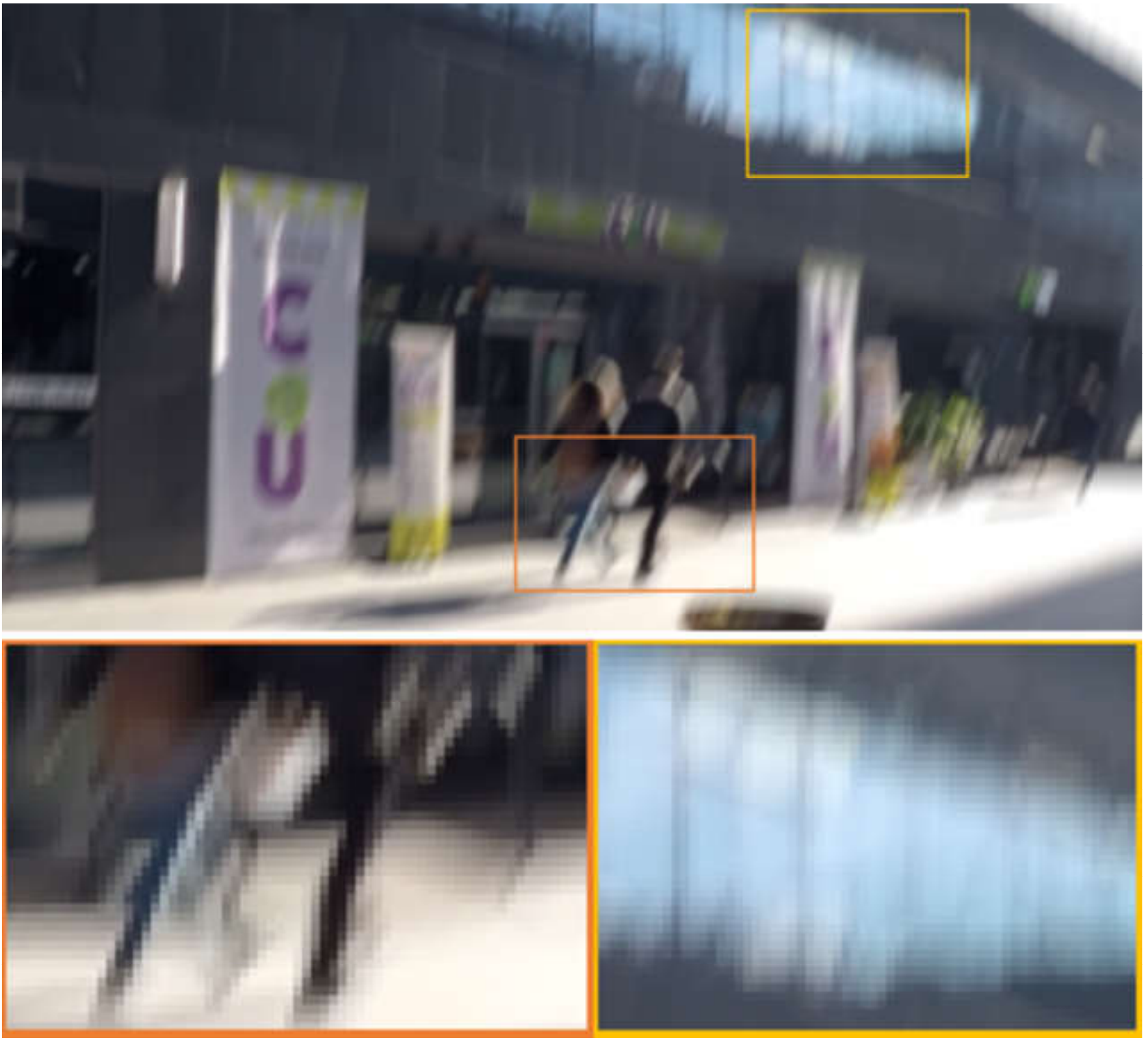} & \hspace{-4mm}
      \includegraphics[width=0.245\linewidth]{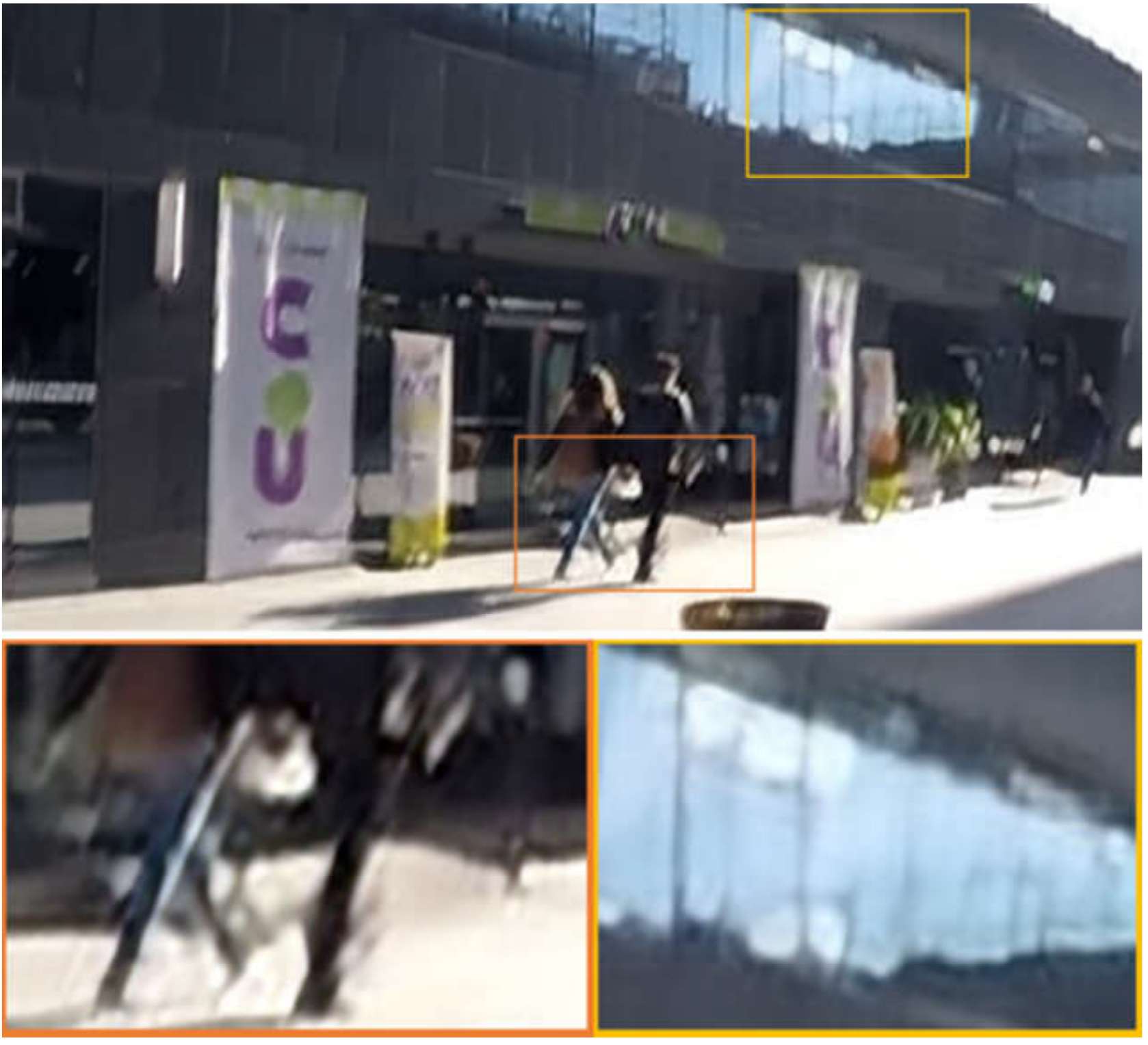} & \hspace{-4mm}
      \includegraphics[width=0.245\linewidth]{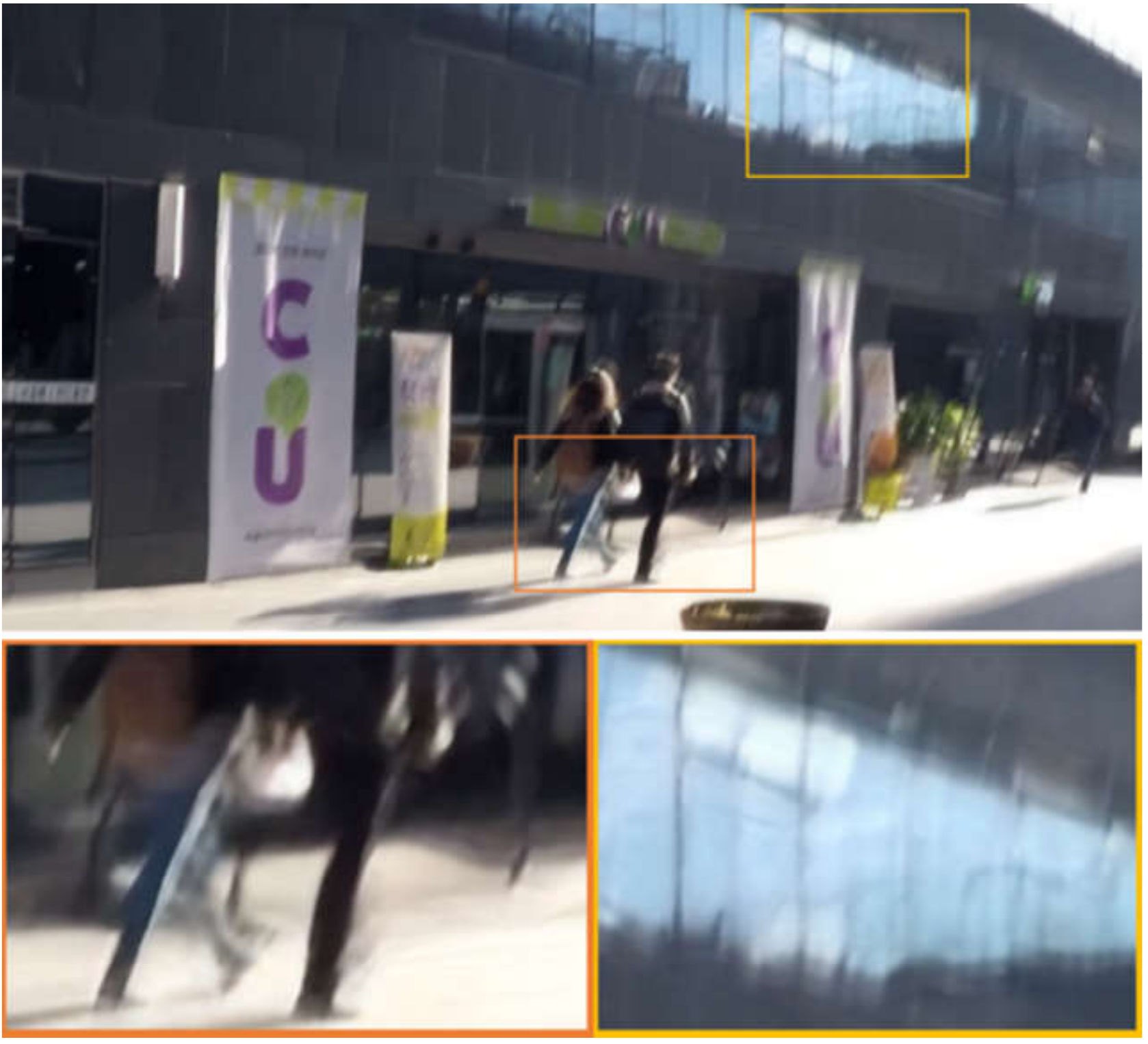}
    \vspace{-0.5mm}
    \\
    \hspace{-3mm} 
      (a) Ground-truth HR & \hspace{-4mm}
    (b) Blurry LR input & \hspace{-4mm}
      (c) DeepDeblur + EDSR & \hspace{-4mm}
      (d) EDSR + DeepDeblur
    \\
    \hspace{-3mm} 
      PSNR / SSIM & \hspace{-4mm}
      21.04 / 0.787 & \hspace{-4mm}
      24.58 / 0.846 & \hspace{-4mm}
      25.04 / 0.876
    \\
    \hspace{-3mm}
      \includegraphics[width=0.245\linewidth]{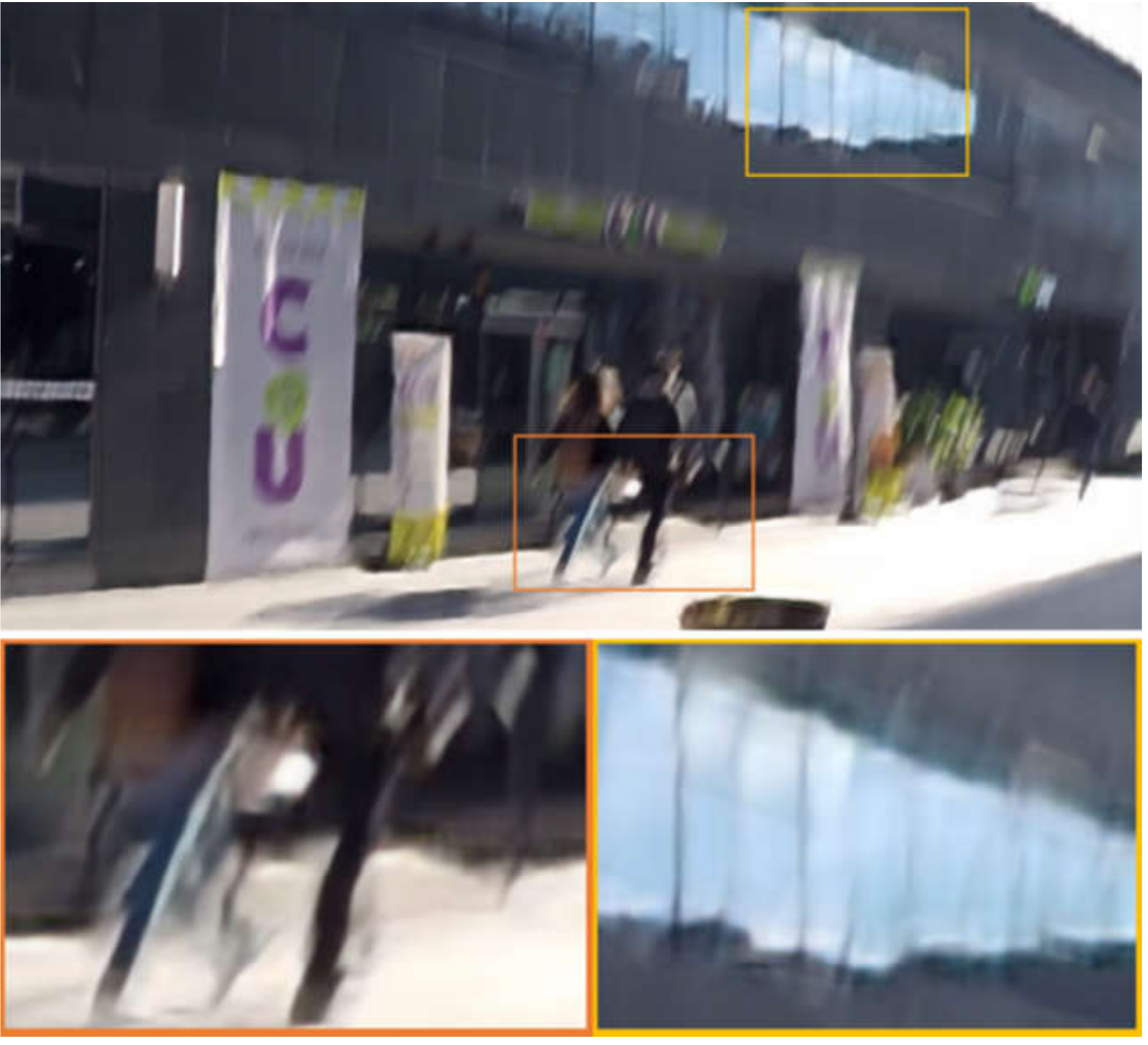} & \hspace{-4mm}
      \includegraphics[width=0.245\linewidth]{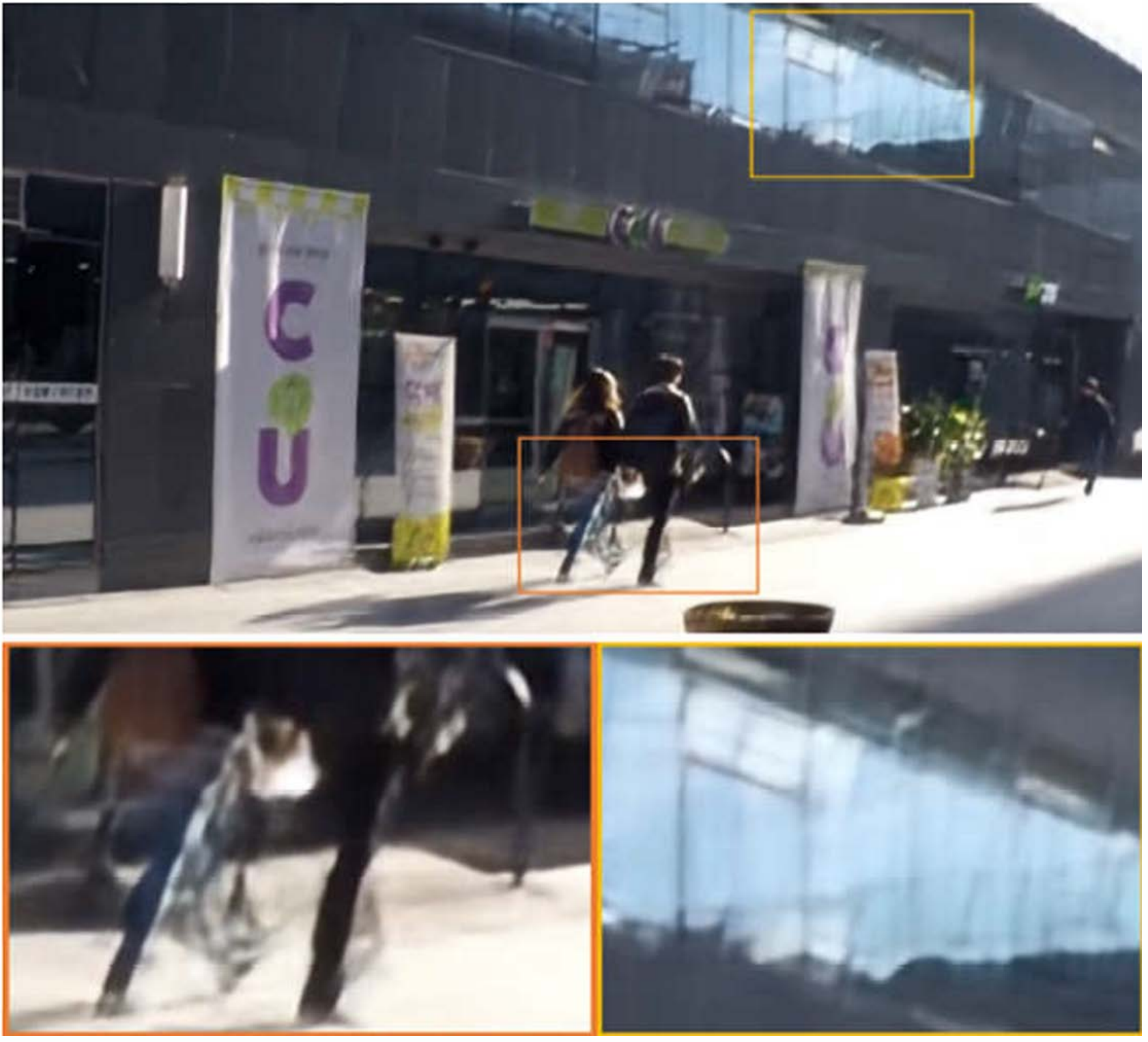} & \hspace{-4mm}
      \includegraphics[width=0.245\linewidth]{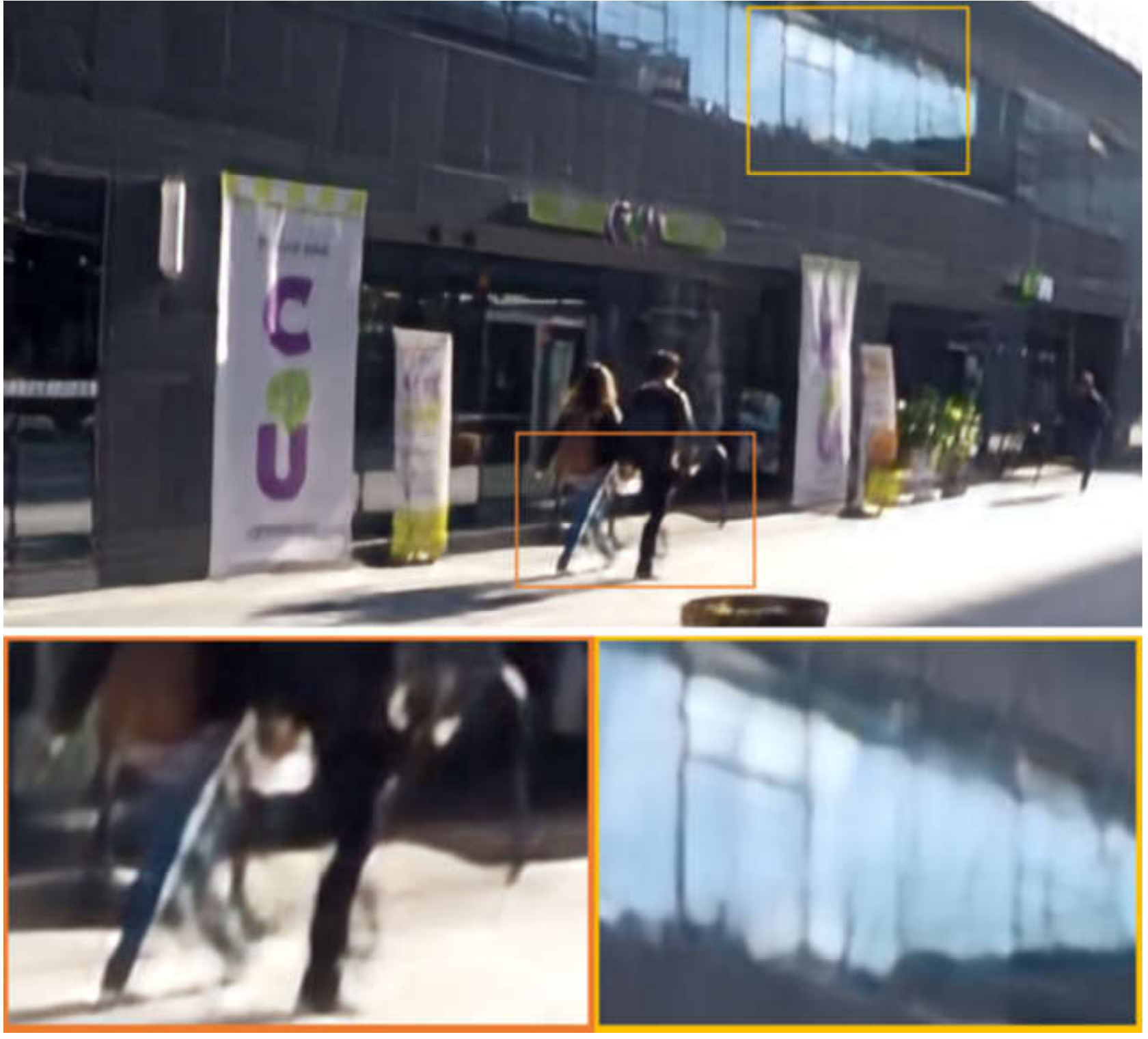} & \hspace{-4mm}
      \includegraphics[width=0.245\linewidth]{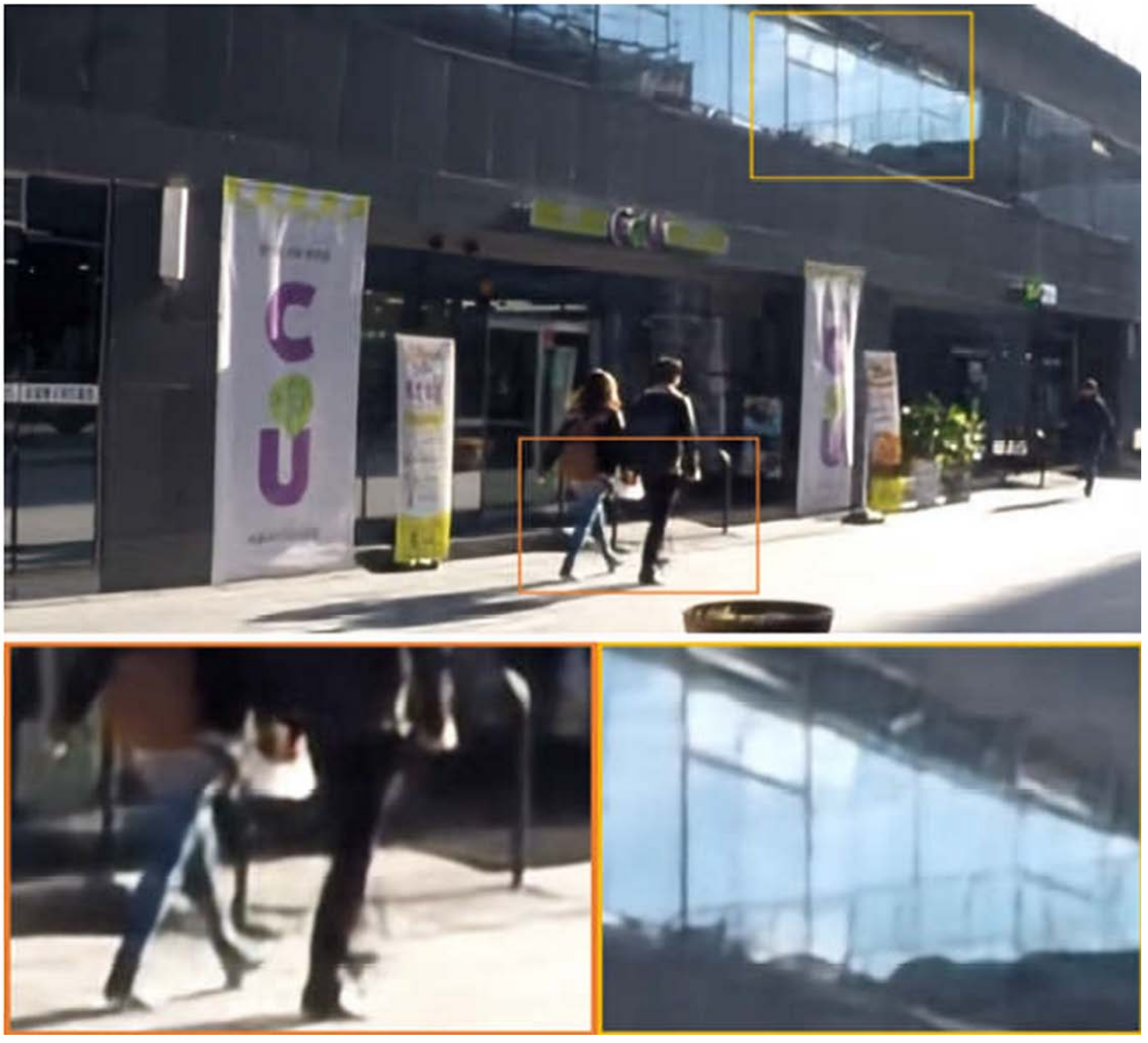}
      \vspace{-0.5mm}
    \\
    \hspace{-3mm} 
      (e) SCGAN${}^{\star}$ & \hspace{-4mm}
      (f) ED-DSRN${}^{\star}$ & \hspace{-4mm} 
      (g) SRResNet${}^{\star}$ & \hspace{-4mm}
      (h) GFN${}^{\star}$ (ours)
    \\
    \hspace{-3mm} 
      23.00 / 0.835 & \hspace{-4mm}
      26.07 / 0.896 & \hspace{-4mm}
      25.63 / 0.881 & \hspace{-4mm}
      29.02 / 0.929
  \end{tabular}
  \caption{
  \textbf{Visual comparison on the LR-GOPRO dataset.}
  The evaluated methods include SRResNet \citep{srresnet}, EDSR \citep{EDSR}, SCGAN \citep{scgan}, ED-DSRN \citep{icassp18}, and DeepDeblur \citep{deepdeblur}.
  The methods with a $\star$ sign are trained on our LR-GOPRO training set.
  The proposed method generates sharper HR images with more details.
  }
  \label{fig:visual_results_GOPRO}
  \end{figure*}

\begin{figure*}[tb]
  \small
  \centering
  \vspace{-1mm}
  \begin{tabular}{cccc}
  \hspace{-3mm}
    \includegraphics[width=0.245\linewidth]{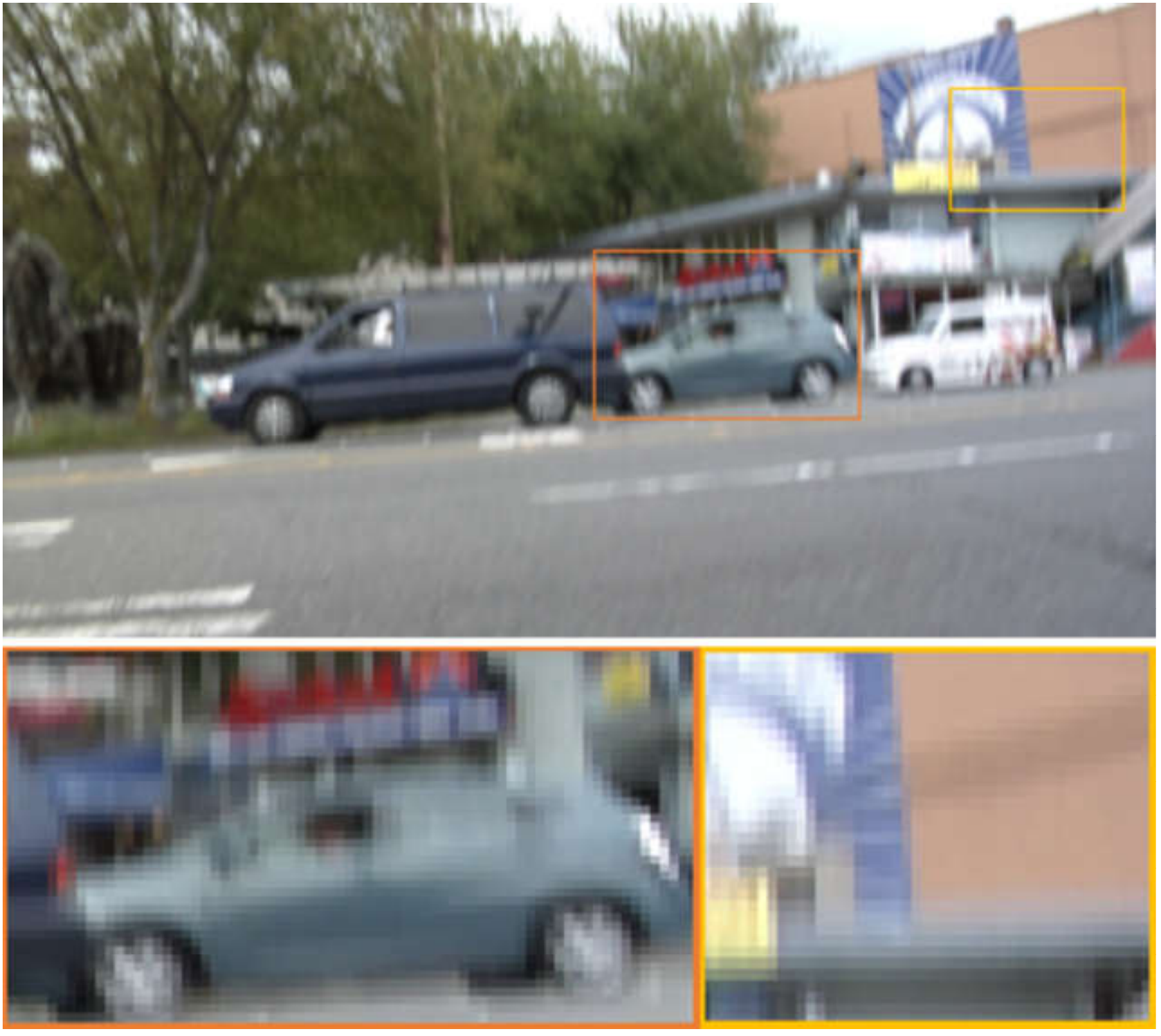} & \hspace{-4mm}
    \includegraphics[width=0.245\linewidth]{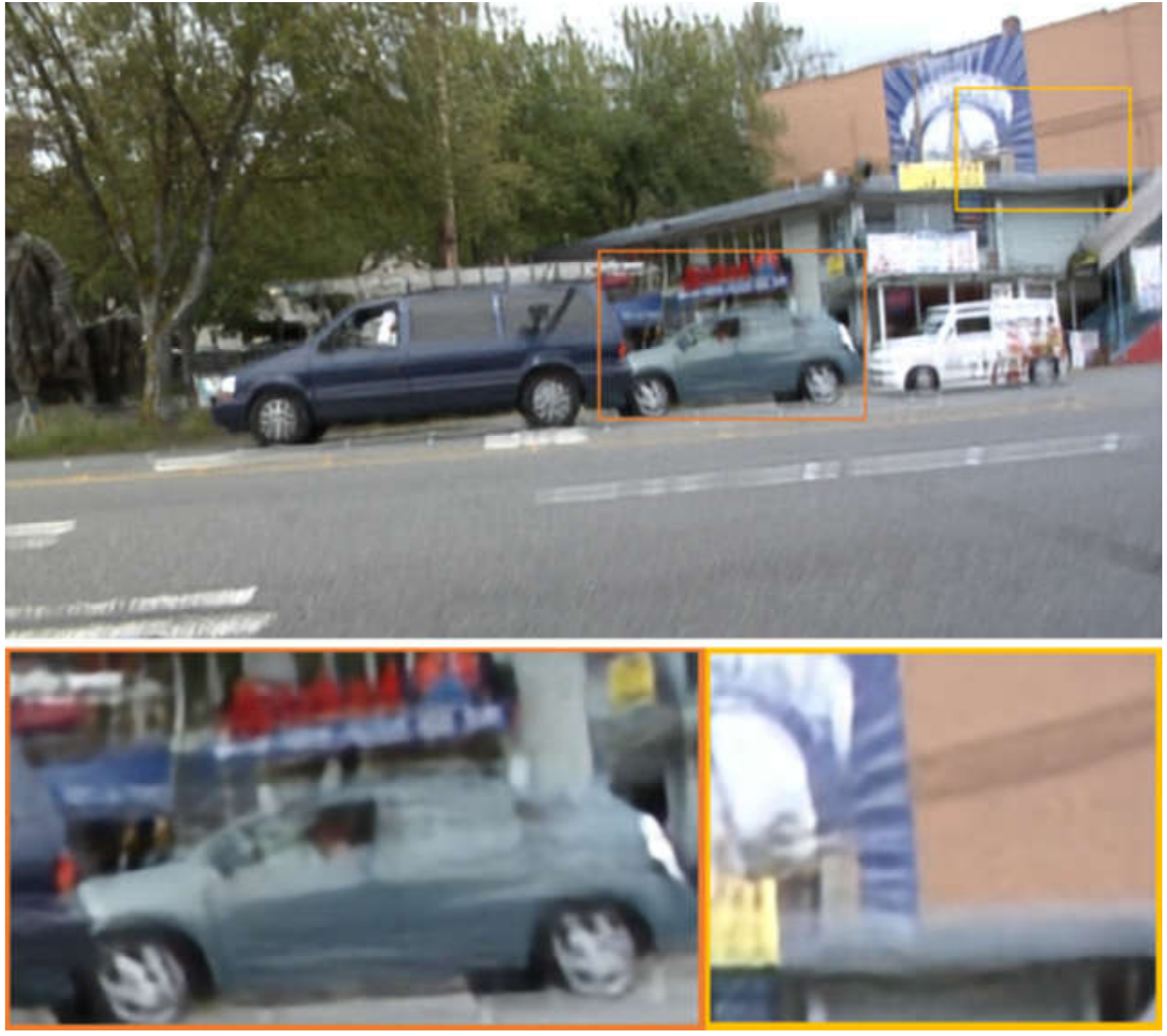} & \hspace{-4mm}
    \includegraphics[width=0.245\linewidth]{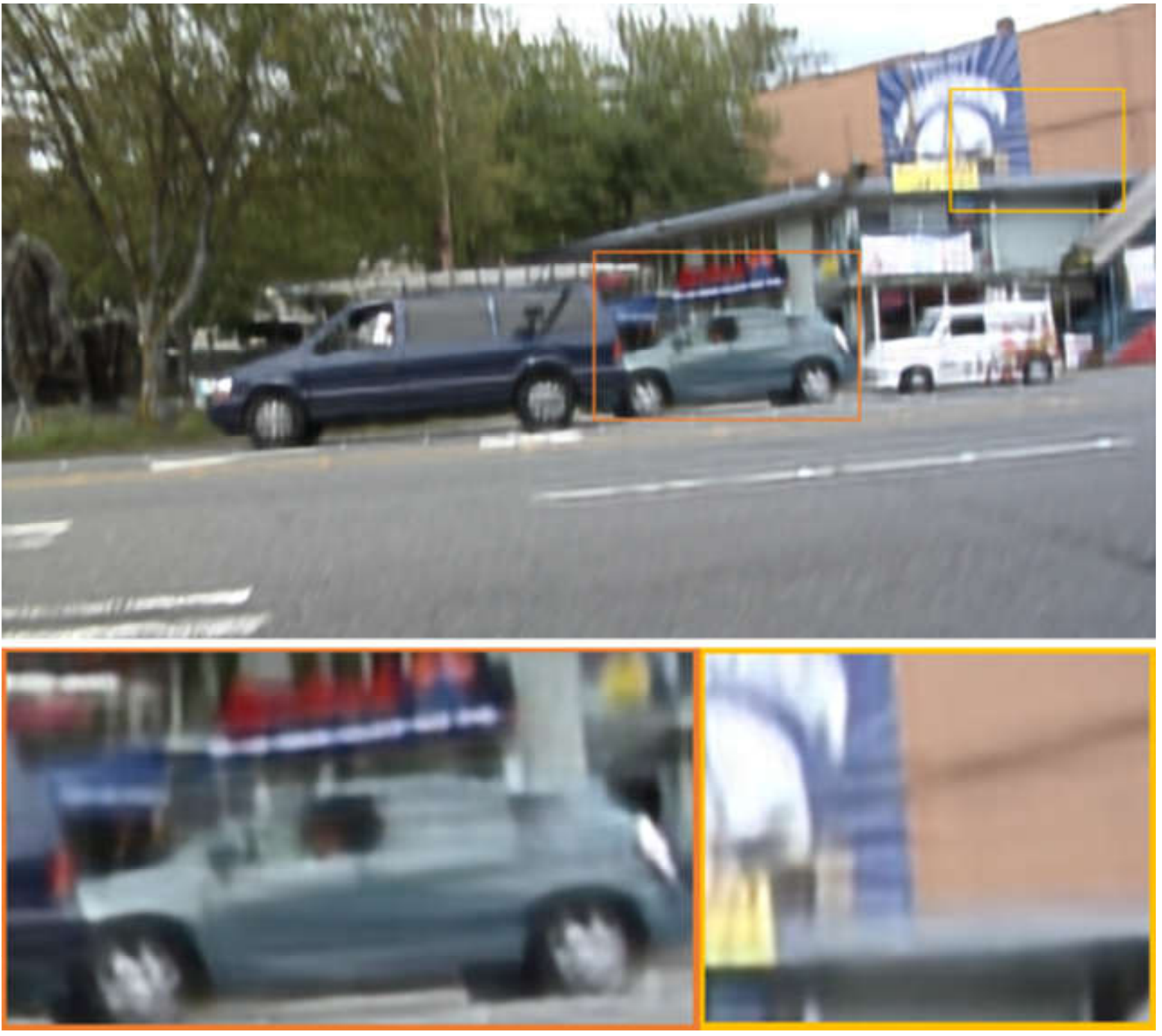} &
  \hspace{-4mm}
    \includegraphics[width=0.245\linewidth]{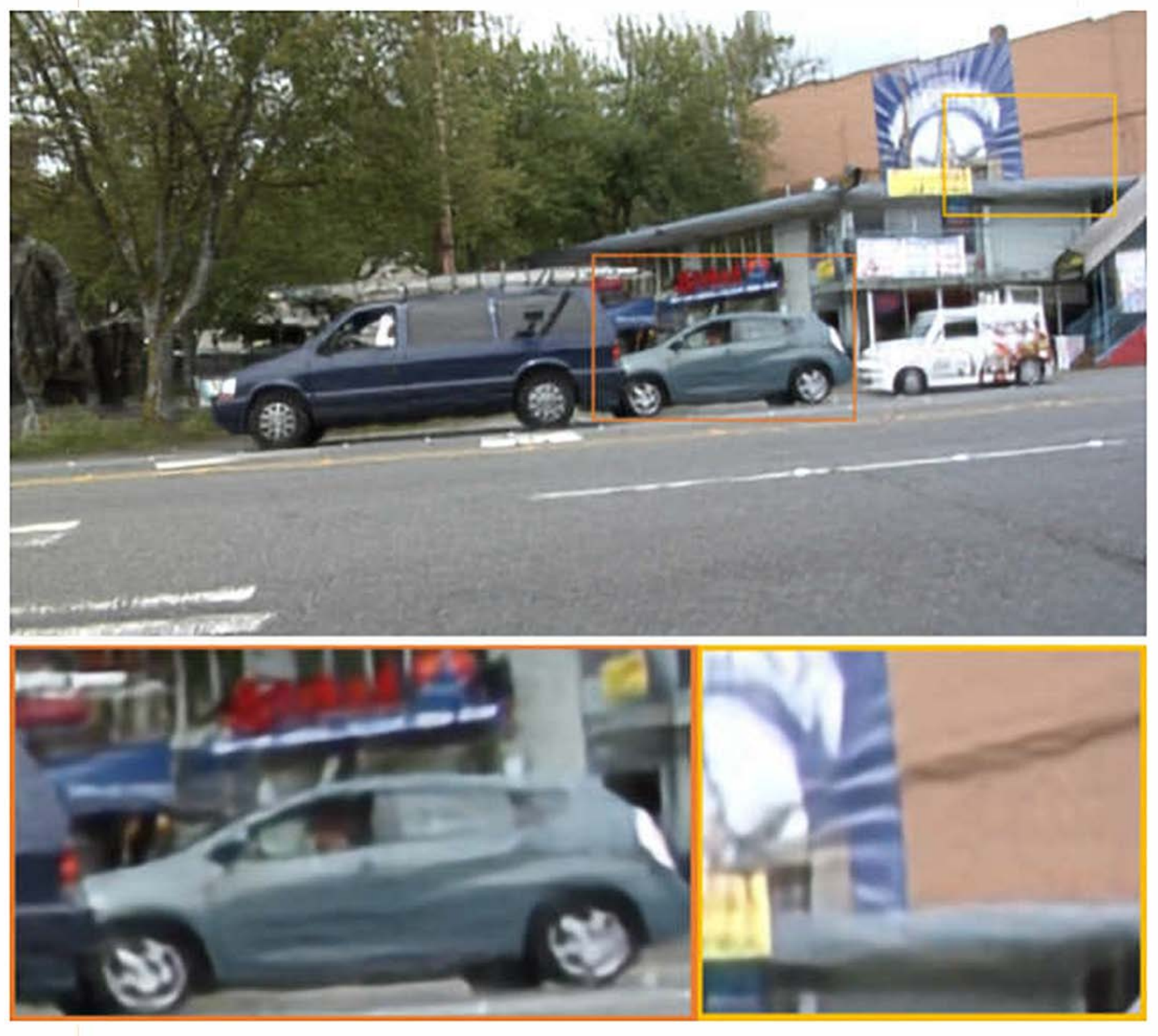}
  \vspace{-0.5mm}
  \\
  \hspace{-4mm} (a) Blurry LR input & (b) ED-DSRN${}^{\star}$ & (c) SRResNet${}^{\star}$ &(d) GFN${}^{\star}$ (ours) 
  \end{tabular}
  \caption{
  \textbf{Visual comparison on the real blurry image dataset \citep{deepvideo}.}
  The methods with a $\star$ sign are trained on our LR-GOPRO training set.
  The proposed GFN model is more robust to outliers in real images and generates sharper results than the re-trained state-of-the-art methods ED-DSRN \citep{icassp18} and SRResNet \citep{srresnet}.
  }
  \label{fig:visual_results_deblur_real}
  \end{figure*}

  We present the qualitative results of the LR-GOPRO dataset in \figref{visual_results_GOPRO} and a real blurry image in \figref{visual_results_deblur_real}.
  The methods using the combination scheme, e.g., DeepDeblur + EDSR and EDSR + DeepDeblur, often introduce undesired artifacts due to the error accumulation problem.
  Existing joint super resolution and deblurring methods (ED-DSRN and SCGAN) do no handle non-uniform blur well.
  In contrast, the proposed algorithm generates sharp HR images with more details.
  
  \subsection{Super Resolving Hazy Image}
  \label{subsec:hazySR}
  \begin{table}[t]
    \small
    \centering
    \caption{
    \textbf{Quantitative comparison with the state-of-the-art methods on super resolving the hazy images.}
    The evaluated methods include \textbf{super resolution methods}, SRResNet \citep{srresnet}, EDSR \citep{EDSR}, RCAN \citep{RCAN},
    \textbf{image dehazing methods}, DCP \citep{dehaze6}, NLD \citep{NLD}, DGFN \citep{GFN}, GCANet \citep{GCANet}, PFFNet \citep{dehaze12}, AODN \citep{AODnet},
    and \textbf{joint approaches}, SCGAN \citep{scgan}, ED-DSRN \citep{icassp18}.
    The methods with a $\star$ sign are trained on our LR-RESIDE training set.
    {\color{red}Red texts} indicate the best performance.
    The proposed GFN performs favorably against existing methods while maintaining a small model size and fast inference speed.
    }
    \label{tab:table_dehaze_results}
    \begin{tabular}{rcl}
    \hline
    &\multicolumn{1}{l}{} & \multicolumn{1}{c}{LR-RESIDE $4\times$}   
    \\
    \multirow{-2}{*}{Method}    &\multirow{-2}{*}{\#Params}     & \multicolumn{1}{c}{PSNR~/~SSIM~/~Time~(s)}    
    \\ 
    \hline
    SRResNet    &1.5M       &13.29~/~0.566~/~{\color{red}0.02}
    \\
    EDSR           &43M      &13.71~/~0.650~/~0.12
    \\
    RCAN         &16M       &13.72~/~0.652~/~0.30
    \\
    SRResNet${}^{\star}$ &1.6M &23.58~/~0.791~/~{\color{red}0.02}
    \\  
    ED-DSRN${}^{\star}$  &25M  &24.89~/~0.813~/~0.04
    \\
    \hline
    DGFN + SRResNet      &2.2M         &19.62~/~0.618~/~0.03
    \\
    SRResNet + DGFN      &2.2M         &19.34~/~0.607~/~0.08
    \\
    AODN + SRResNet    &1.5M        &16.49~/~0.597~/~{\color{red}0.02}
    \\
    SRResNet + AODN   &1.5M        &17.07~/~0.563~/~0.05
    \\
    EDSR + DCP         &16M       &17.46~/~0.572~/~24.0
    \\
    EDSR + NLD         &16M       &17.70~/~0.576~/~10.6
    \\ 
    AODN + EDSR        &43M      &17.05~/~0.702~/~0.12
    \\
    EDSR + AODN       &43M        &18.30~/~0.713~/~0.13
    \\ 
    DGFN + EDSR          &44M        &21.03~/~0.740~/~0.13   
    \\
    EDSR + DGFN          &44M        &21.89~/~0.775~/~0.18
    \\ 
    DGFN + RCAN          &16M        &21.04~/~0.740~/~0.30
    \\
    RCAN + DGFN         &16M       &21.92~/~0.777~/~0.36
    \\ 
    PFFNet + RCAN       &38M        &20.55~/~0.678~/~0.31
    \\
    RCAN + PFFNet       &38M       &23.76~/~0.795~/~0.31
    \\
    RCAN + GCANet         &17M       &22.93~/~0.786~/~1.2
    \\ 
    \hline
    GFN${}^{\star}$~(ours)                  &12M         &{\color{red}25.77}~/~{\color{red}0.830}~/~{\color{red}0.02}~
    \\
    \hline
    \end{tabular}
    \end{table}
    
  {\flushleft \bf Training Dataset and Details.}
  We use the RESIDE \citep{RESIDE} dataset to generate the training data for the joint super resolution and dehazing problem.
  For training, we randomly select 5005 outdoor hazy and sharp HR image pairs in 35 different haze concentrations and 5000 indoor HR pairs in 10 different haze concentrations from RESIDE training sets.
  We apply the same procedure as the LR-GOPRO dataset to generate the training triplets of $\left\{L_{haze}, L, H\right\}$.
  We refer to the generated dataset as LR-RESIDE in the following.
  
\begin{figure*}[tb]
    \small
    \centering
    \begin{tabular}{cccc}
      \hspace{-3mm}
       \includegraphics[width=0.245\linewidth]{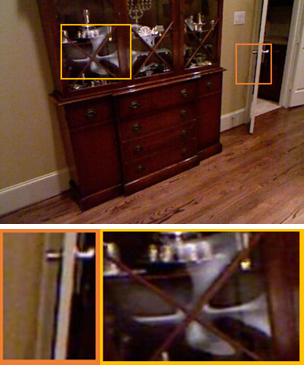} & \hspace{-4mm}
        \includegraphics[width=0.245\linewidth]{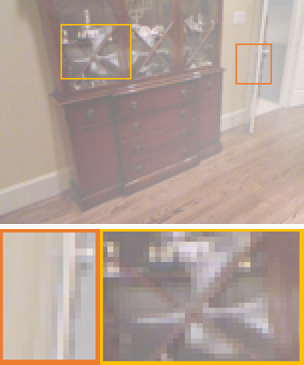} & \hspace{-4mm}
        \includegraphics[width=0.245\linewidth]{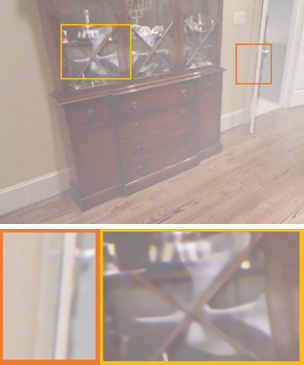} & \hspace{-4mm}
        \includegraphics[width=0.245\linewidth]{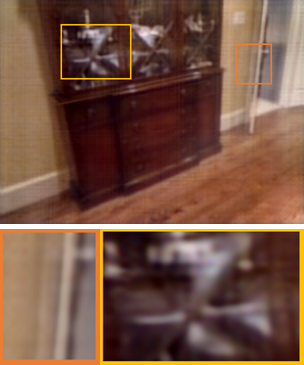}
      \vspace{-0.5mm}
      \\
      \hspace{-3mm} 
        (a) Ground-truth HR & \hspace{-4mm}
        (b) Hazy LR input & \hspace{-4mm}
        (c) RCAN  & \hspace{-4mm}
        (d) PFFNet + RCAN 
      \\
      \hspace{-3mm} 
        PSNR / SSIM & \hspace{-4mm}
        7.28 / 0.401 & \hspace{-4mm}
        7.30 / 0.419 & \hspace{-4mm}
        17.08 / 0.800
      \\
      \hspace{-3mm}
        \includegraphics[width=0.245\linewidth]{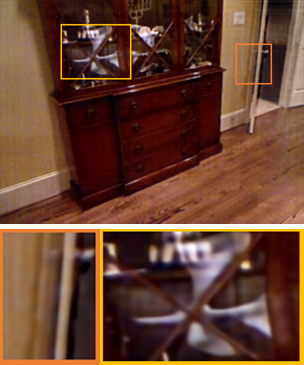} & \hspace{-4mm}
        \includegraphics[width=0.245\linewidth]{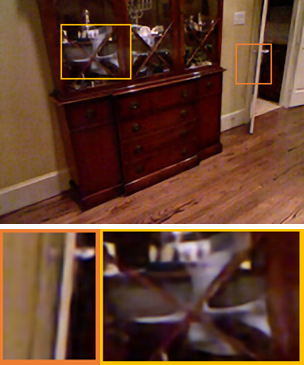} & \hspace{-4mm}
        \includegraphics[width=0.245\linewidth]{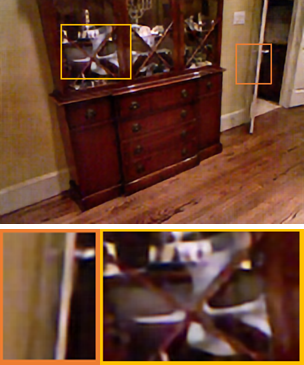} & \hspace{-4mm}
        \includegraphics[width=0.245\linewidth]{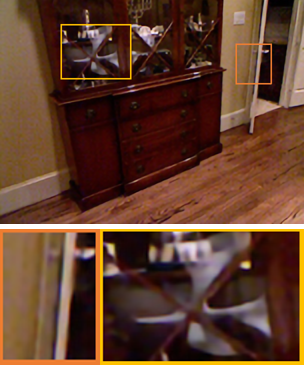}
        \vspace{-0.5mm}
      \\
      \hspace{-3mm} 
        (a) RCAN + PFFNet & \hspace{-4mm}
        (b) ED-DSRN${}^{\star}$ & \hspace{-4mm}
        (c) SRResNet${}^{\star}$ & \hspace{-4mm}
        (d) GFN${}^{\star}$ (ours)
      \\
      \hspace{-3mm} 
        24.03 / 0.920 & \hspace{-4mm}
        27.23 / 0.952 & \hspace{-4mm}
        24.84 / 0.917 & \hspace{-4mm}
        28.85 / 0.961
    \end{tabular}
    \caption{
    \textbf{Visual comparison on the LR-RESIDE dataset.}
    The evaluated methods include SRResNet \citep{srresnet}, RCAN \citep{RCAN}, ED-DSRN \citep{icassp18}, and PFFNet \citep{dehaze12}.
    The methods with a $\star$ sign are trained on our LR-RESIDE training set.
    The proposed model generates sharper HR images with more details.
    }
    \label{fig:visual_results_RESIDE}
    \end{figure*}

  Since the training process of joint dehazing and super resolution is more stable compared with that of joint deblurring and super resolution, we simplify the training process into two stages. 
  First, we train the network without the gate module from scratch for 25 epochs.
  The learning rate is set to $10^{-4}$ and multiplied by 0.5 for every 7 epochs.
  Second, we enable the gate module and train the complete model for 60 epochs where the learning rate is set to $10^{-4}$ and multiplied by 0.1 for every 25 epochs.
  The other settings are the same as those for blurry image super resolution.
  %
  {\flushleft \bf Performance Evaluation.}
  %
    We choose 500 indoor image pairs and 500 outdoor image pairs from the test set of the RESIDE dataset for evaluation.
  We compare the proposed GFN model with the state-of-the-art super resolution methods \citep{srresnet,EDSR,RCAN}, joint image deblurring and super resolution approaches \citep{icassp18, scgan}, and combinations of super resolution algorithms and dehazing schemes \citep{dehaze6, NLD, GFN,dehaze12,AODnet, GCANet}.
  For fair comparisons, we re-train the models of SRResNet \citep{srresnet}, ED-DSRN \citep{icassp18}, and PFFNet \citep{dehaze12} on our training set\footnote{Since the pre-trained model of the PFFNet is not available, we train the network directly on the RESIDE dataset and achieve  quantitative results on the RESIDE dataset better than the reported results. We use this model in the following experiments.}.
  Other super resolution methods are trained on the DIV2K dataset \citep{DIV2K} 
  and deep learning-based dehazing methods are trained on the RESIDE dataset \citep{RESIDE}.
 \\ 
  The quantitative evaluations in \tabref{table_dehaze_results} show that the proposed GFN model performs well in terms of PSNR and SSIM with shorter inference time.
  We present qualitative results on the LR-RESIDE dataset in \figref{visual_results_RESIDE}.
  The state-of-the-art super resolution method (RCAN) does not remove the haze from the hazy input, and the straightforward combination schemes, e.g., PFFNet + RCAN and RCAN + PFFNet, generate undesired artifacts and distorted colors on the flat regions due to the error accumulation problem.
  The re-trained SRResNet and ED-DSRN methods do not recover the details well.
  In contrast, the proposed model generates better results with more details.

    \begin{table}[t]
    \small
    \centering
    \caption{
    \textbf{Quantitative comparison with the state-of-the-art methods on super resolving the rainy images.}
    The evaluated methods include \textbf{super resolution methods}, SRResNet \citep{srresnet}, EDSR \citep{EDSR}, RCAN \citep{RCAN}, 
    \textbf{image deraining methods}, RESN \citep{RESCAN}, IDGAN \citep{ID-CGAN}, DID-MDN \citep{DID-MDN},
    and \textbf{joint approaches}, SCGAN \citep{scgan}, ED-DSRN \citep{icassp18}.
    The methods with a $\star$ sign are trained on our LR-Rain1200 training set.
    {\color{red}Red texts} indicate the best performance.
    The GFN-Large scheme performs favorably against existing methods while maintaining a small model size and fast inference speed.
    }
    \label{tab:table_results_deraining}
    \begin{tabular}{rcl}
    \hline
    &\multicolumn{1}{l}{} & \multicolumn{1}{c}{LR-Rain1200 $4\times$}  
    \\
    \multirow{-2}{*}{Method}    &\multirow{-2}{*}{\#Params}     & \multicolumn{1}{c}{PSNR~/~SSIM~/~Time~(s)}      
    \\ 
    \hline
    SRResNet   &1.5M         &16.27~/~0.341~/~0.02~ 
    \\
    EDSR           &43M        &19.88~/~0.548~/~0.12
    \\
    RCAN          &16M         &19.94~/~0.548~/~0.30
    \\
    SRResNet${}^{\star}$ &1.6M         &23.29~/~0.624~/~{\color{red}0.02}~         
    \\  
    ED-DSRN${}^{\star}$ &25M     &23.86~/~0.694~/~0.04~   
    \\
    \hline
    RESN  + SRResNet &1.7M    &19.01~/~0.497~/~0.05~
    \\
    SRResNet + RESN   &1.7M   &17.75~/~0.386~/~0.72~
    \\
    IDGAN + SRResNet &1.8M    &18.28~/~0.451~/~0.30~
    \\
    SRResNet + IDGAN &1.8M    &17.25~/~0.407~/~0.37~
    \\ 
    RESN  + EDSR &43M    &20.71~/~0.617~/~0.15~
    \\
    EDSR + RESN &43M    &22.37~/~0.644~/~0.82~
    \\
    IDGAN + EDSR &43M    &19.57~/~0.581~/~0.40~ 
    \\
    EDSR + IDGAN &43M    &19.96~/~0.606~/~0.47~ 
    \\ 
    RESN + RCAN     &16M  &20.74~/~0.618~/~0.33~                          
    \\
    RCAN  + RESN      &16M  &22.53~/~0.650~/~1.00~
    \\ 
    DID-MDN + RCAN      &82M   &22.31~/~0.610~/~0.31~  
    \\
    RCAN  + DID-MDN     &82M   &23.50~/~0.685~/~0.35~ 
    \\ 
    \hline
    GFN${}^{\star}$~(ours)      &12M        &24.64~/~0.705~/~0.02~
    \\
    GFN-Large${}^{\star}$~(ours)               &24M        &{\color{red}25.24}~/~{\color{red}0.709}~/~{\color{red}0.02}~
    \\
    \hline
    \end{tabular}
    \end{table}

  \subsection{Super Resolving Rainy Image}
  \label{subsec:rainySR}
\begin{figure}[tb]
    \centering
    \begin{tabular}{cc}
      \includegraphics[width=0.40\linewidth]{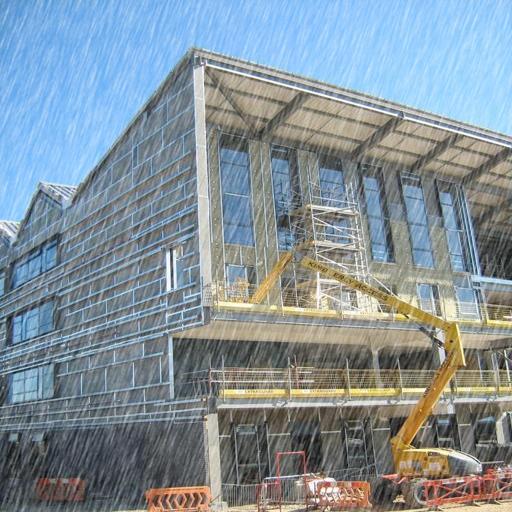} & \hspace{-2mm}
      \includegraphics[width=0.40\linewidth]{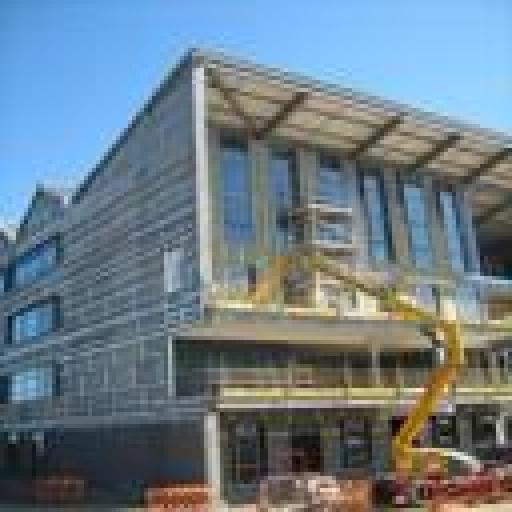} \\
      (a)  & (b) \\
      \includegraphics[width=0.40\linewidth]{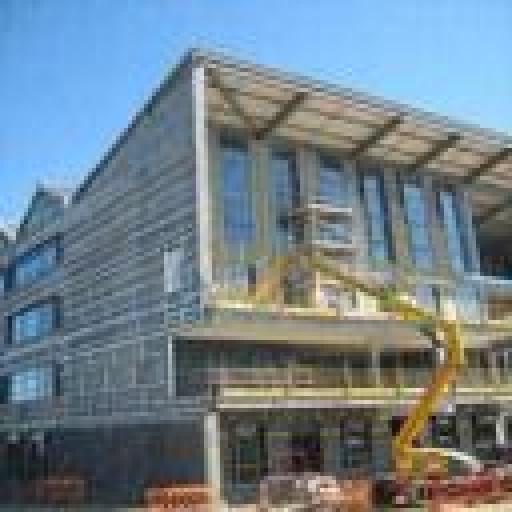} & \hspace{-2mm}
      \includegraphics[width=0.40\linewidth]{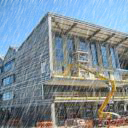} \\
     (c) & (d) \\
    \end{tabular}
    \caption{
    \textbf{Examples of the generated low-resolution image for super resolving rainy images.}
    To generate training and testing dataset for super resolving rainy images, we use rainy/sharp image pairs from Rain1200 dataset \citep{DID-MDN} as the HR images. 
    Simply applying bicubic downsampling on the rainy HR image (a) results in the LR image (c), where many rain streaks are removed. 
    Thus, we first obtain a sharp LR image (b) by applying bicubic downsampling on the sharp HR image and generate the rainy LR image (d) by synthesizing rain streaks using the approach in \citep{DID-MDN}.
    }
    \label{fig:rainy_datasets}
    \end{figure}

  {\flushleft \bf Training Dataset and Details.}
  Since there is no off-the-shelf dataset for rainy image super resolution, we use the Rain1200 \citep{DID-MDN} dataset to generate rainy LR images.
  We note that directly applying  bicubic downsampling on the rainy HR images tends to remove rain streaks as this operator, similar to low-pass filtering, reduces high-frequency details such as thin structures.
  As shown in \figref{rainy_datasets}, the LR image directly downsampled from rainy HR image does not contain many rain streaks and is similar to the LR image downsampled from the clean HR image (see \figref{rainy_datasets}(b) and (c)).
  In order to obtain more realistic LR inputs, we generate rainy LR images by synthesizing rainy streaks on downsampled sharp images.
  We first apply bicubic downsampling on the sharp HR image $H$ to generate the sharp LR image $L$. 
  Similar to \citep{DID-MDN}, we use Photoshop to synthesize rain streaks on $L$ to generate the rainy LR image $L_{rain}$.
  After data augmentation, we obtain 24,000 triplets of $\left\{L_{rain}, L, H\right\}$ for training and 1,200 triplets for testing.
  We refer to the generated dataset as LR-Rain1200 in the following.
  
  To remove long streaks in the rainy images, we modify the network structure of the restoration module to enlarge the receptive field.
  Specifically, we use the structure in \citep{dehaze12} as the encoder-decoder architecture of the restoration module.
  Since the resolution of a rainy LR input in this dataset is relatively low ($128 \times 128$), we remove the last strided convolutional layer and set the output channels of the rest three scales to 64, 128, and 256 respectively. 
  Moreover, we apply the residual learning scheme in the reconstruction module $G_{recon}$ to accelerate the training process.
  We use a deconvolutional layer with the filter size of $4 \times 4$ to upsample the rainy LR input before merging with the output of the reconstruction module $G_{recon}$.
  The training processes are mostly the same as the one for the blurry image super resolution task except that we use a batch size of 6 due to limited GPU memory.

\begin{figure*}[tb]
    \small
    \centering
    \begin{tabular}{cccc}
      \hspace{-3mm}
       \includegraphics[width=0.245\linewidth]{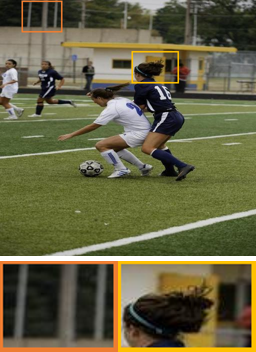} & \hspace{-4mm}
        \includegraphics[width=0.245\linewidth]{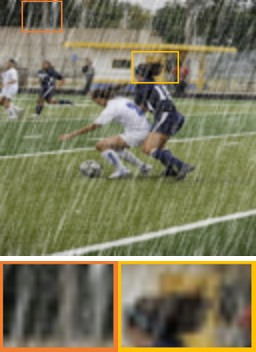} & \hspace{-4mm}
        \includegraphics[width=0.245\linewidth]{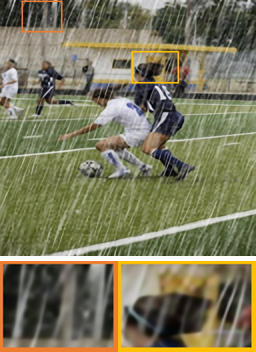} & \hspace{-4mm}
        \includegraphics[width=0.245\linewidth]{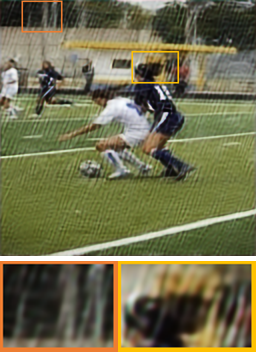}
      \vspace{-0.5mm}
      \\
      \hspace{-3mm} 
        (a) Ground-truth HR & \hspace{-4mm}
        (b) Rainy LR input & \hspace{-4mm}
        (c) RCAN  & \hspace{-4mm}
        (d) DID-MDN + RCAN 
      \\
      \hspace{-3mm} 
        PSNR / SSIM & \hspace{-4mm}
        18.80 / 0.761 & \hspace{-4mm}
        18.58 / 0.726 & \hspace{-4mm}
        22.28 / 0.806
      \\
      \hspace{-3mm}
        \includegraphics[width=0.245\linewidth]{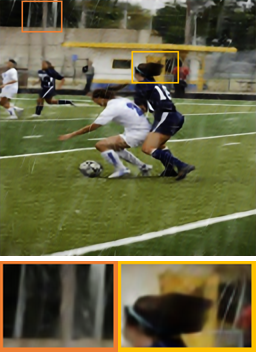} & \hspace{-4mm}
        \includegraphics[width=0.245\linewidth]{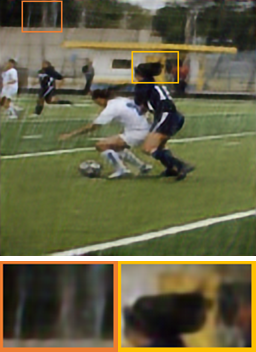} & \hspace{-4mm}
        \includegraphics[width=0.245\linewidth]{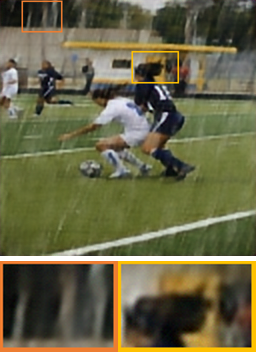} & \hspace{-4mm}
        \includegraphics[width=0.245\linewidth]{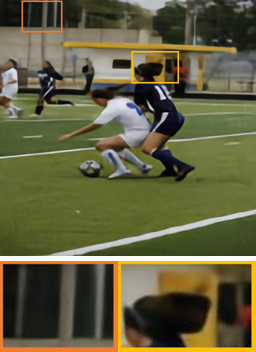}
        \vspace{-0.5mm}
      \\
      \hspace{-3mm} 
        (a) RCAN + DID-MDN & \hspace{-4mm}
        (b) ED-DSRN${}^{\star}$ & \hspace{-4mm}
        (c) SRResNet${}^{\star}$ & \hspace{-4mm}
        (d) GFN-Large${}^{\star}$ (ours)
      \\
      \hspace{-3mm} 
        24.90 / 0.872 & \hspace{-4mm}
        24.62 / 0.828 & \hspace{-4mm}
        24.10 / 0.826 & \hspace{-4mm}
        25.50 / 0.869
    \end{tabular}
    \caption{
    \textbf{Visual comparison on the LR-Rain1200 dataset.}
    The evaluated methods include SRResNet \citep{srresnet}, RCAN \citep{RCAN}, ED-DSRN \citep{icassp18}, and DID-MDN \citep{DID-MDN}.
    The methods with a $\star$ sign are trained on our LR-Rain1200 training set.
    The proposed model is able to remove rain streaks and generates sharper HR images with more details.
    }
    \label{fig:visual_results_Rain1200}
    \end{figure*}

  {\flushleft \bf Performance Evaluation.}
   \tabref{table_results_deraining} shows the quantitative results in terms of PSNR, SSIM, and average inference time.
   Since there exists no approach for joint image deraining and super resolution, we evaluate against the state-of-the-art super resolution algorithms \citep{srresnet,EDSR,RCAN}, joint image deblurring and super resolution approaches \citep{icassp18,scgan}, and straightforward combinations of super resolution and deraining schemes \citep{DID-MDN,ID-CGAN,RESCAN}.
   For fair comparisons, we re-train the SRResNet \citep{srresnet} and ED-DSRN \citep{icassp18} models on our training set.
   The other super resolution methods are trained on the DIV2K dataset \citep{DIV2K} 
   and deep learning-based deraining methods are trained on the Rain800 dataset \citep{ID-CGAN} (RESN and IDGAN) and Rain1200 dataset \citep{DID-MDN} (DID-MDN).
   As shown in \tabref{table_results_deraining}, the proposed model with a large receptive field, denoted by GFN-Large, achieves better performance with shorter inference time and fewer model parameters than the evaluated methods.
   Some deblurred results are shown in \figref{visual_results_Rain1200}.
   Although the RCAN method recovers some high-frequency details, it does not remove the rain streaks on the image.
   The re-trained SRResNet model and straightforward combination approaches, DID-MDN + RCAN and RCAN + DID-MDN, do not remove long rain streaks and often introduce unexpected artifacts on the rich texture regions due to the error accumulation problem.
   The re-trained ED-DSRN model removes most rain streaks but does not restore clear contours and high-frequency details.
   The proposed GFN-Large model accurately removes the rain streaks while preserving the structural information and recovering more details.

\subsection{Ablation Study and Analysis}
\label{subsec:Ablation}
\begin{figure*}[htb]
  \centering
  \begin{tabular}{cc}
  \hspace{-4mm}
    \includegraphics[width=0.45\linewidth]{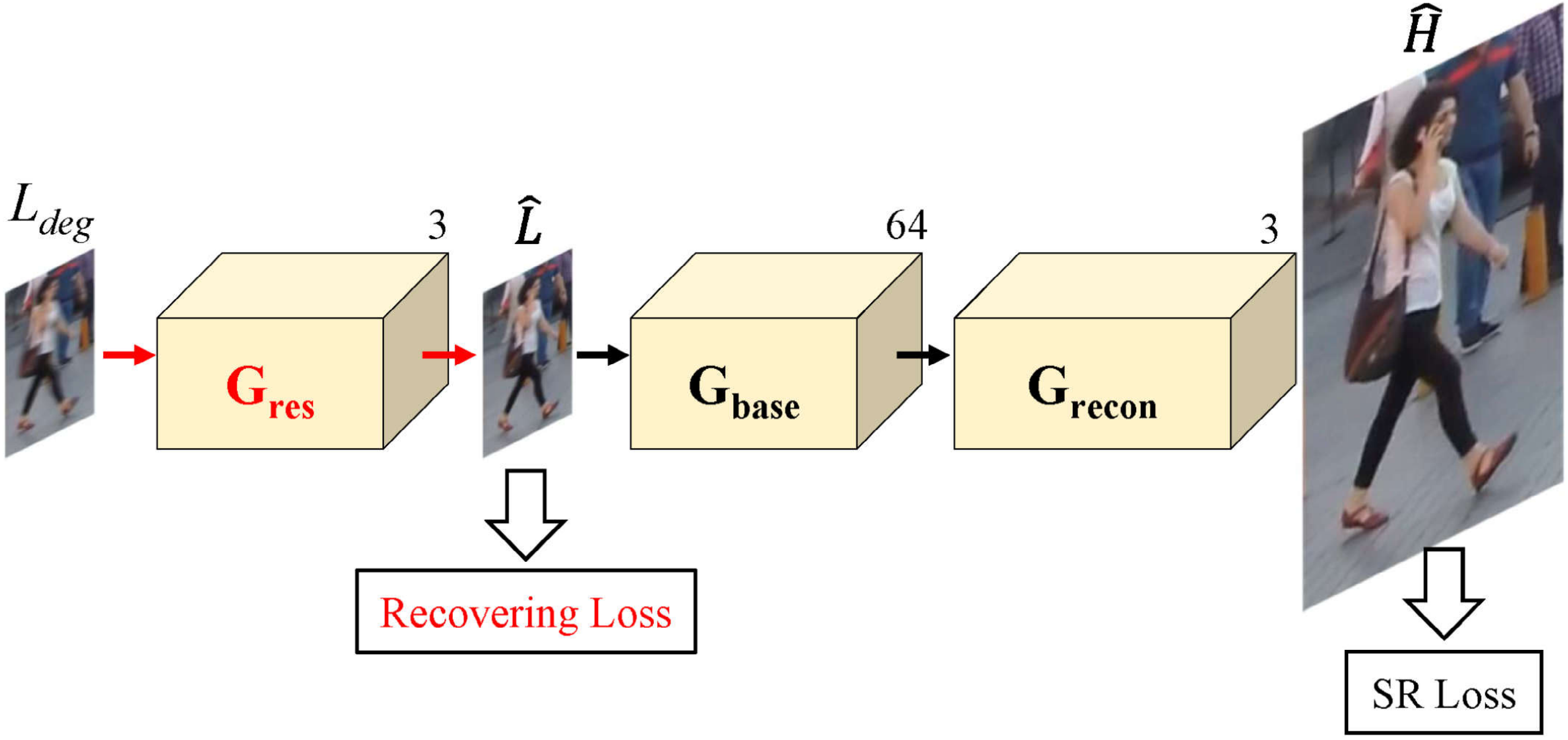} & \hspace{2mm}
    \includegraphics[width=0.45\linewidth]{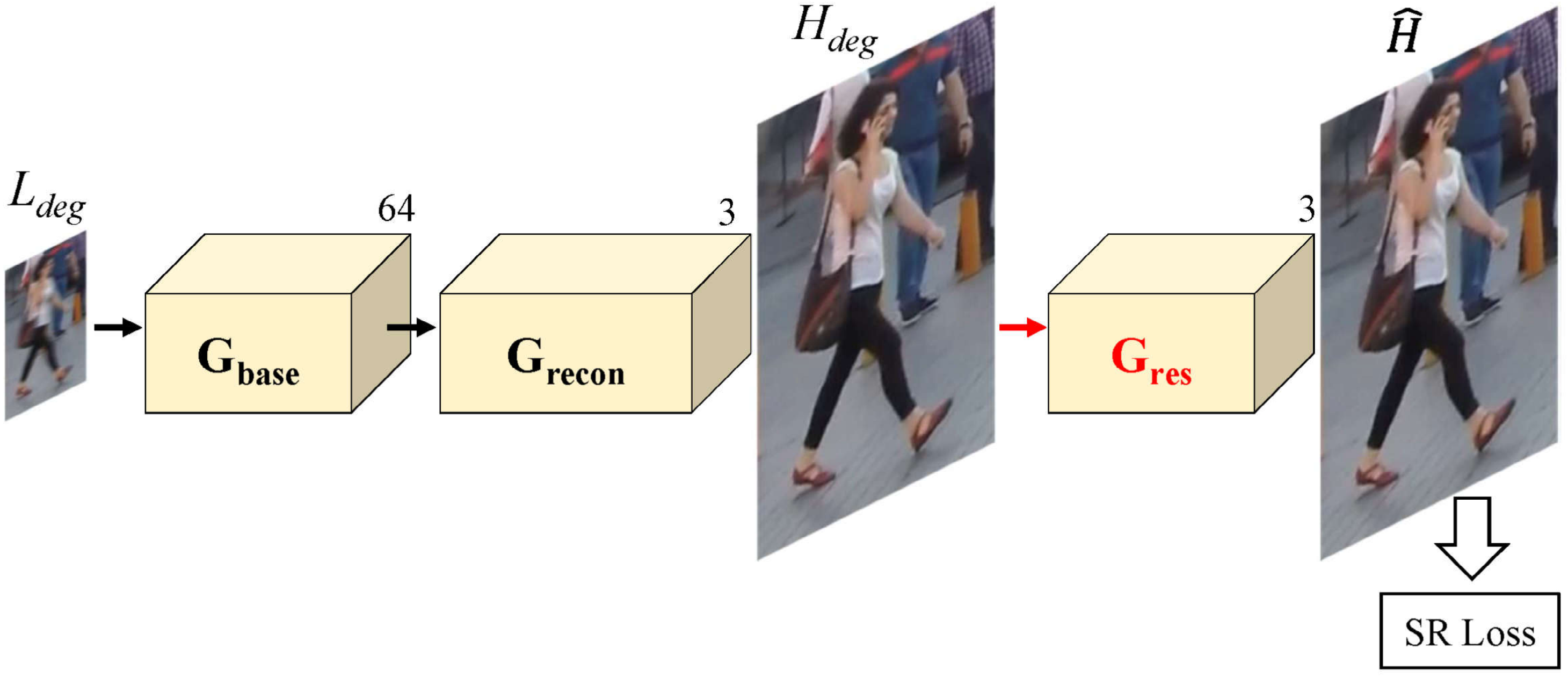}
  \\
  \hspace{-4mm}(a) Model-1 & \hspace{-2mm} (b) Model-2
  \\
  \vspace{1mm} & \vspace{1mm}
  \\
  \hspace{-4mm}
    \includegraphics[width=0.45\linewidth]{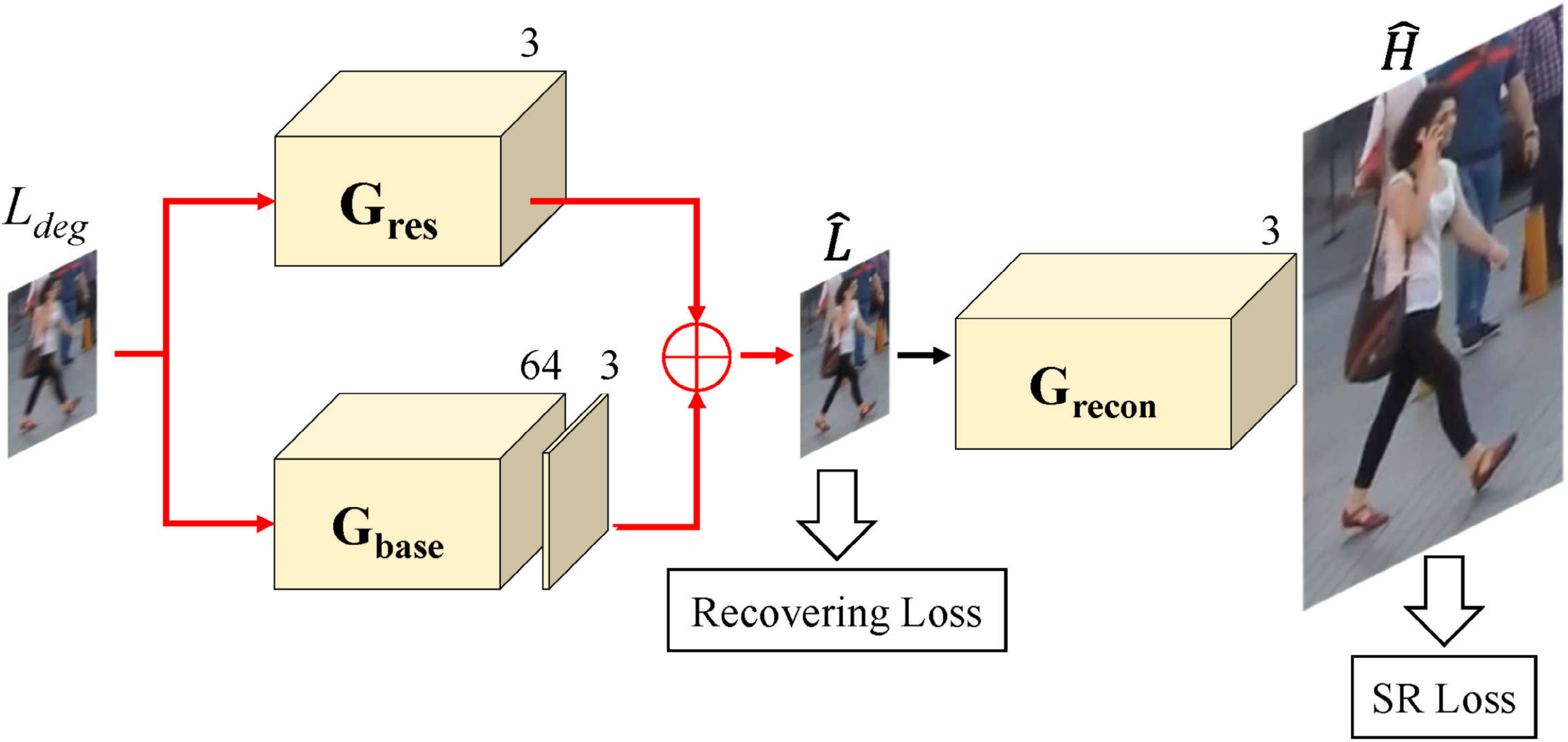} & \hspace{2mm}
    \includegraphics[width=0.45\linewidth]{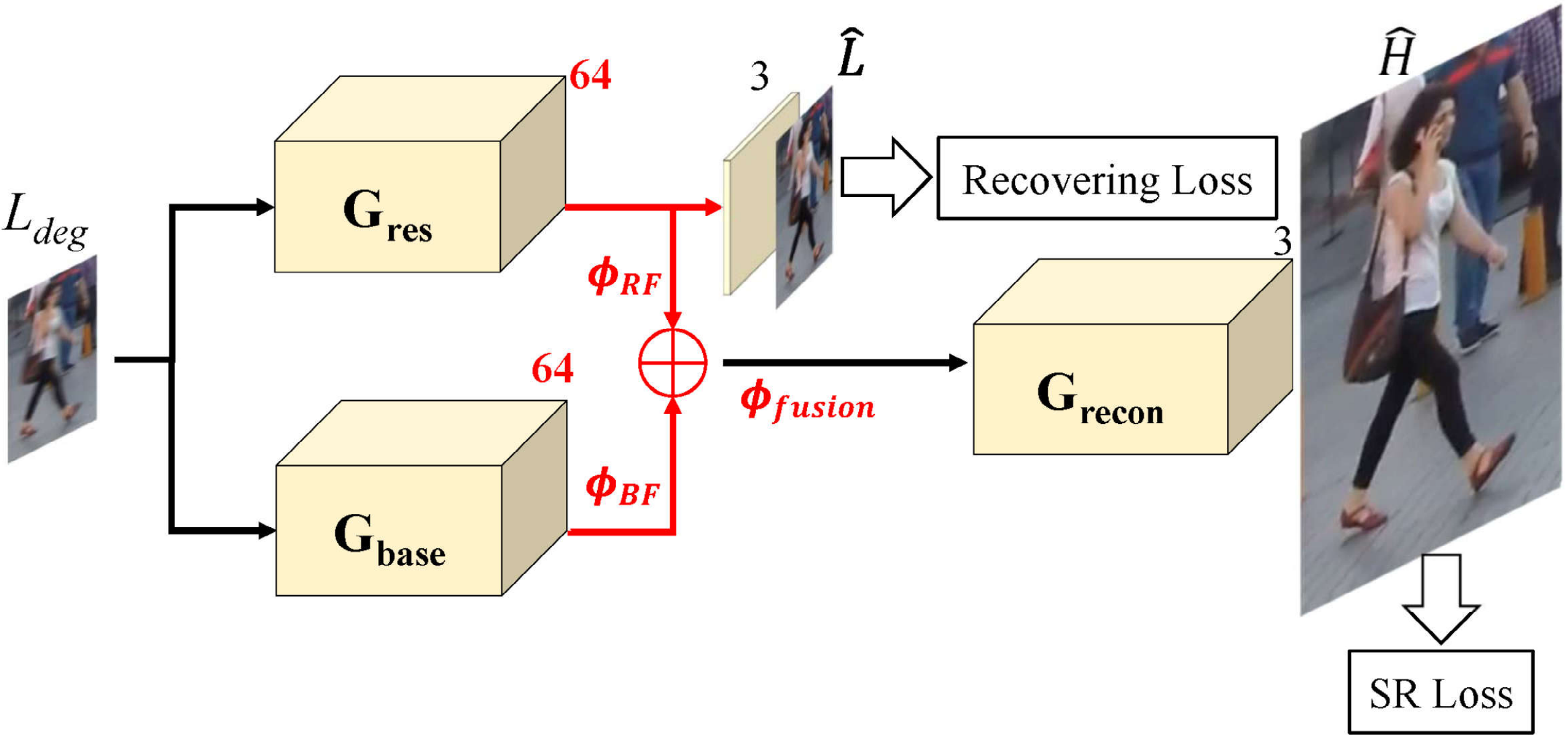}
  \\
  \hspace{-4mm} (c) Model-3 & \hspace{-2mm} (d) Model-4
  \\
  \vspace{1mm} & \vspace{1mm}
  \end{tabular}
  \caption{
  \textbf{Network structure of the models in the ablation study.} $G_{res}$, $G_{base}$,  $G_{recon}$ represent the restoration module, 
  base feature extraction module, and reconstruction module in GFN.}
  \label{fig:supp_ablation}
  \end{figure*}

The proposed GFN consists of four modules:
a \textnormal{restoration module} to extract recovered features;
a \textnormal{dual-branch} architecture instead of a concatenation of base and restoration modules;
a fusion approach on the \textnormal{feature level};
and a \textnormal{gate module} to adaptively fuse base and recovered features.
To further analyze the components, we train the combination of the base feature extraction module and reconstruction module ($G_{base}+G_{recon}$) as the baseline 
and introduce other modules progressively to evaluate them.
All the models in this section are trained from scratch using the same settings for fair comparisons.
Without loss of generality, we conduct these experiments on two applications, blurry image super resolution and hazy image super resolution.
The evaluated network architectures are illustrated in Figure \ref{fig:supp_ablation}, and the results are shown in \tabref{ablation}. 

%
{\flushleft \bf Effect of Restoration Module.}
We use the restoration module and baseline model in two ways: restoration first (Model-1) and SR first (Model-2).
The restoration module shows significant performance improvement over the baseline on both applications (0.62 dB and 1.24 dB for blurry and hazy images, respectively).
The SR-first combination scheme achieves better performance but has slower execution speed.

{\flushleft \bf Effect of Dual-Branch Architecture.}
We use a dual-branch structure to extract the base and recovered features separately (Model-3 listed in \tabref{ablation}).
The outputs of the two modules are fused by direct addition, and the recovering loss is used to guide this process.
Compared with the sequential restoration and super resolution method (Model-1), it achieves 0.19 dB and 0.28 dB improvement on blurry and hazy images, respectively.
Furthermore, the Model-3 performs comparably with the SR-first method (Model-2) but more efficiently.
This is due to the heavy computation load of the restoration process carried out in the HR feature space for Model-2.

%
{\flushleft \bf Effect of Feature Level Fusion.}
Since the recovering loss in the Model-3 is computed after fusion, it does not provide explicit guidance on each branch.
In the Model-4, we impose the recovering loss on the restoration branch as explicit regularization.
Furthermore, to reduce the computational redundancy and avoid error accumulation, we fuse the branches on the feature level, instead of fusing them on the pixel level.
Compared with the Model-3, the Model-4 achieves 0.52 dB and 0.11 dB performance improvement and lower computational cost on two tasks.
{\flushleft \bf Effect of Gate Module.}
We introduce the gate module to enable local and channel-wise feature fusion from two branches.
The gate module also helps exploit the dependence between the features. 
Here, we evaluate the gate module in terms of the number of recursive blocks $N$.
As shown in \tabref{ablation}, the gate module with 3 recursive blocks performs best, with improvements of 0.38 dB and 0.58 dB over the Model-4 on two tasks. 
We note that the gate module with more than $3$ recursive blocks does not perform well.

{\flushleft \bf Generalizability of GFN.}
To show that the proposed architecture is a generic framework, we replace the original restoration and reconstruction modules with more advanced network architectures and show that it can obtain further performance gains.
  We use the Residual Dense Block (RDB) in the RDN \citep{RDN} to replace the ResBlock in the reconstruction module and use the dilation architecture in the GCANet \citep{GCANet} to replace the classical encoder-decoder architecture in the restoration module. 
  As shown in \tabref{new_module}, using the more advanced structure improves the performance under the same training settings.
More ablation study are included in the appendix.
    
    \begin{table*}[]
      \small
      \centering
      \caption{
      \textbf{Analysis of key components in the proposed GFN.}
      All models are trained on the LR-GOPRO dataset or the LR-RESIDE dataset with the same hyper-parameters.
      The baseline method is the network with the base feature extraction module and the reconstruction module.
      DB is the abbreviation for ``deblurring'', and DH is the abbreviation for ``dehazing''.
      }
      \label{table_Ablation}
      \begin{tabular*}{1\textwidth}{@{\extracolsep{\fill}}r|p{1.10cm}<{\centering}p{1.15cm}<{\centering} p{1.15cm}<{\centering} p{1.15cm}<{\centering} p{1.15cm}<{\centering}|p{1.2cm}<{\centering}p{1.2cm}<{\centering}p{1.2cm}<{\centering}p{1.2cm}<{\centering}}
      \hline
      \multirow{-1}{*}{Modifications}  &Baseline &Model-1  &Model-2 &Model-3 &Model-4 &$GFN_{N=1}$   &$GFN_{N=2}$  &$GFN_{N=3}$ &$GFN_{N=4}$
      \\ 
      \hline
      restoration module       &	&\checkmark	&\checkmark	&\checkmark	&\checkmark	&\checkmark     &\checkmark   &\checkmark &\checkmark 
      \\
      dual-branch            &	&	&	&\checkmark	&\checkmark	&\checkmark &\checkmark 
      &\checkmark &\checkmark
      \\
      feature level    &	&	&	&	&\checkmark	&\checkmark	  &\checkmark &\checkmark  &\checkmark
      \\
      gate module             &	&	&	&	&	&\checkmark	         &\checkmark &\checkmark &\checkmark
      \\
      SR-first                &	&	&\checkmark 	&	&	&	  \\
      \hline
       DB + SR PSNR(dB)                    &26.20	&26.82	&27.00	&27.01		&27.53  &27.74	&27.87 &27.91 &27.86	
      \\
      Time (s)                &0.07	&0.09	&0.57	&0.10		&0.07	&0.07   &0.07   &0.07 &0.07
      \\
      \hline
      DH + SR PSNR(dB) &23.58 &24.82 &25.12 &25.10 &25.21 &25.44 &25.69 &25.77 &25.72
      \\
      Time (s) &0.02 &0.03 &0.16 &0.03 &0.02 &0.02 &0.02 &0.02 &0.02 
      \\
      \hline
      \end{tabular*}
      \label{tab:ablation}
      \vspace{-2mm}
      \end{table*}

  \begin{table*}[tb]
        \small
        \centering
        \caption{
          \textbf{Qualitative results using different restoration and reconstruction modules on the LR-RESIDE dataset.}
          We evaluate the following methods: replacing the ResBlock in the proposed GFN with the Residual Dense Block (RDB) in the RDN \citep{RDN} (GFN\_RDN), replacing the restoration module in the proposed GFN with the dilation architecture in the GCANet \citep{GCANet} (GFN\_GCANet), and replacing both of them (GFN\_RDN\_GCANet). 
          All the models are trained using the same setting.
        }
        \begin{tabular}[t]{c|cccc}
        \hline
        \multirow{-1}{*}{}  &GFN     &GFN\_RDN &GFN\_GCANet &GFN\_RDN\_GCANet
        \\ 
        \hline
        PSNR/SSIM                  &25.77/0.830	&25.88/0.833	&25.83/0.831	 &25.90/0.833	
        \\
        \hline
        \end{tabular}
        \label{tab:new_module}
        \vspace{-2mm}
  \end{table*}

  \subsection{Limitations}
  \vspace{-3mm}
  To remove non-local degradation, such as haze or long rain streaks, we use an encoder-decoder architecture to extract global and contextual information.
  However, this approach does not effectively extract local features commonly used for super resolution \citep{EDSR, srresnet}.
  As a result, the proposed method tends to generate over-smoothed results compared to the other schemes, as shown in some regions of \figref{visual_results_RESIDE} and \figref{visual_results_Rain1200}. 
  For future work, we will explore more effective architectures to better exploit both global and local visual information for super resolution on degraded images.
\vspace{-3mm}

  \section{Conclusions}
  In this paper, we propose an end-to-end architecture to recover a sharp HR image from a degraded LR input.
  The proposed network consists of two branches to extract recovered and base features separately.
  The extracted features are fused through a recursive gate module and  
  used to reconstruct the final results.
  The network design decouples the joint problem into two restoration tasks and enables efficient training and inference.
  Extensive evaluations of different restoration tasks demonstrate that the proposed model is effective for super resolving degraded images.

\section*{Acknowledgement}
This work is partially supported by National Major Science and Technology Projects of China grant under number 2019ZX01008103, 
National Natural Science Foundation of China (61603291), 
Natural Science Basic Research Plan in Shaanxi Province of China (Program No.2018JM6057), 
and the Fundamental Research Funds for the Central Universities. 
%
  

\bibliographystyle{spbasic}

\clearpage

\section*{Appendix}
\renewcommand\thesubsection{\Alph{subsection}} 
\setcounter{table}{0} 
\renewcommand{\thetable}{\Alph{table}}
\setcounter{figure}{0} 
\renewcommand{\thefigure}{\Alph{figure}}
\label{sec:supp}

\subsection{\textbf{Network Configuration}} 
\label{subsec:supp-A}
We present the detailed configuration of the proposed network in \tabref{supp_table_network}, with respect to the four modules in the network: 
the \textbf{deblurring module}, \textbf{SR feature extraction module}, \textbf{recursive gate module}, and \textbf{reconstruction module}.

\begin{table*}[!hp]
  \fontsize{6.8}{11}\selectfont
  \newcommand{\tabincell}[2]{\begin{tabular}{@{}#1@{}}#2\end{tabular}}
  \centering
  \caption{\textbf{Configuration of the proposed network.} The values in the \textbf{skip} row are \textbf{layer} names, indicating whose outputs are added to the outputs of the corresponding layers.}
  \label{tab:supp_table_network}
  \hspace{-0mm}
  \begin{tabular}[t]{|c|c|c|c|c|}
  \hline
  \multicolumn{5}{|c|}{\textbf{Restoration Module}}
  \\
  \hline
  \textbf{layer} 
  &\textbf{output size}
  &\textbf{kernel}
  &\textbf{LReLU}
  &\textbf{skip}
  \\
  \hline
  Input\_1 
  &$3\times h \times w$
  &
  &
  &
  \\
  \hline
  conv1 
  &$64\times h \times w$
  &7
  &
  &
  \\
  \hline
  \tabincell{c}{Resblock \\ 1-6}
  &$64\times h \times w$
  &3
  &
  &conv1
  \\
  \hline
  conv2
  &$128\times \frac{h}{2} \times \frac{w}{2}$
  &3
  &
  &
  \\
  \hline
  \tabincell{c}{Resblock \\ 7-12}
  &$128\times \frac{h}{2} \times \frac{w}{2}$
  &3
  &
  &conv2
  \\
  \hline
  conv3
  &$256\times \frac{h}{4} \times \frac{w}{4}$
  &3
  &
  &
  \\
  \hline
  \tabincell{c}{Resblock \\ 13-18}
  &$256\times \frac{h}{4} \times \frac{w}{4}$
  &3
  &
  &conv3
  \\
  \hline
  deconv1
  &$128\times \frac{h}{2} \times \frac{w}{2}$
  &4
  &\checkmark
  &
  \\
  \hline
  deconv2
  &$64\times h \times w$
  &4
  &\checkmark
  &
  \\
  \hline
  conv4
  &$64\times h \times w$
  &7
  &
  &conv1
  \\
  \hline
  conv5
  &$64\times h \times w$
  &3
  &\checkmark
  &
  \\
  \hline
  conv6
  &$3\times h \times w$
  &3
  &
  &
  \\
  \hline
  \multicolumn{5}{|c|}{\textbf{Base Feature Extraction Module}}
  \\
  \hline
  \textbf{layer} 
  &\textbf{output size}
  &\textbf{kernel}
  &\textbf{LReLU}
  &\textbf{skip}
  \\
  \hline
  Input\_1 
  &$3\times h \times w$
  &
  &
  &
  \\
  \hline
  conv7 
  &$64\times h \times w$
  &7
  &
  &
  \\
  \hline
  \tabincell{c}{Resblock \\ 19-26}
  &$64\times h \times w$
  &3
  &
  &
  \\
  \hline
  conv8 
  &$64\times h \times w$
  &3
  &
  &conv7
  \\
  \hline
  \end{tabular}
  \qquad
  \hspace{-5mm}
  \begin{tabular}[t]{|c|c|c|c|c|}
  \hline
  \multicolumn{5}{|c|}{\textbf{Gate Module}}
  \\
  \hline
  Input\_2.0
  &$131\times h \times w$
  &
  &
  &
  \\
  \hline
  conv9 
  &$64\times h \times w$
  &3
  &\checkmark
  &
  \\
  \hline
  conv10 
  &$64\times h \times w$
  &1
  &
  &
  \\
  \hline
  \tabincell{c}{Elementwise \\ mul} 
  &$64\times h \times w$
  &
  &
  &conv8
  \\
  \hline
  Input\_2.1
  &$131\times h \times w$
  &
  &
  &
  \\
  \hline
  conv9 
  &$64\times h \times w$
  &3
  &\checkmark
  &
  \\
  \hline
  conv10 
  &$64\times h \times w$
  &1
  &
  &
  \\
  \hline
  \tabincell{c}{Elementwise \\ mul} 
  &$64\times h \times w$
  &
  &
  &input\_2.1
  \\
  \hline
  Input\_2.2
  &$131\times h \times w$
  &
  &
  &
  \\
  \hline
  conv9 
  &$64\times h \times w$
  &3
  &\checkmark
  &
  \\
  \hline
  conv10 
  &$64\times h \times w$
  &1
  &
  &
  \\
  \hline
  \tabincell{c}{Elementwise \\ mul} 
  &$64\times h \times w$
  &
  &
  &input\_2.2
  \\
  \hline
  \multicolumn{5}{|c|}{\textbf{Reconstruction Module}}
  \\
  \hline
  Input\_3
  &$64\times h \times w$
  &
  &
  &
  \\
  \hline
  \tabincell{c}{Resblock \\ 27-34}  
  &$64\times h \times w$
  &3
  &
  &
  \\
  \hline
  conv11 
  &$256\times h \times w$
  &3
  &
  &
  \\
  \hline
  \tabincell{c}{pixel \\ shuffle}  
  &$64\times 2h \times 2w$
  &
  &\checkmark
  &
  \\
  \hline
  conv12 
  &$256\times 2h \times 2w$
  &3
  &
  &
  \\
  \hline
  \tabincell{c}{pixel \\ shuffle}  
  &$64\times 4h \times 4w$
  &
  &\checkmark
  &
  \\
  \hline
  conv13 
  &$64\times 4h \times 4w$
  &3
  &\checkmark
  &
  \\
  \hline
  conv14 
  &$3\times 4h \times 4w$
  &3
  &
  &
  \\
  \hline
  \end{tabular}
  \end{table*}

\subsection{\textbf{List of the Evaluated Methods}} 
\label{subsec:supp-B}
All the the evaluated methods in Section \ref{sec:experiment} are listed in \tabref{list}.

\setcounter{table}{1}
\begin{table*}[b]
  \small
  \centering
  \caption{
  \textbf{List of the evaluated methods in Section 4.}
  }
  \label{table_Citation}
  \begin{adjustbox}{width=0.99\linewidth}
  \begin{tabular}[t]{cc}
  \hline
  Method  & Reference 
  \\ 
  \hline
  SRResNet   & MSE-based SRResNet in ``Photo-Realistic Single Image Super-Resolution Using a Generative Adversarial Network'' by \cite{srresnet}	
  \vspace{0.5mm}
  \\
  EDSR   & EDSR in ``Enhanced deep residual networks for single image super-resolution'' by \cite{EDSR}                
  \vspace{0.5mm}
  \\
  RDN   & RDN (D=20, C=6, and G=32) in ``Residual dense network for image super-resolution'' by \cite{RDN}                
  \vspace{0.5mm}
  \\
  RCAN   & ``Image super-resolution using very deep residual channel attention networks'' by \cite{RCAN}                
  \vspace{0.5mm}
  \\
  \hline
  SCGAN & MSE-based SCGAN in ``Learning to super-resolve blurry face and text images'' by \cite{scgan}
  \vspace{0.5mm}
  \\
  ED-DSRN & ``A deep encoder-decoder networks for joint deblurring and super-resolution'' by \cite{icassp18}
  \vspace{0.5mm}
  \\
  \hline
  DeepDeblur & ``Deep multi-scale convolutional neural network for dynamic scene deblurring'' by \cite{deepdeblur}
  \vspace{0.5mm}
  \\
  DeblurGAN & ``DeblurGAN: Blind motion deblurring using conditional adversarial networks'' by \cite{deblurgan}
  \vspace{0.5mm}
  \\
  SRN & ``Scale-recurrent network for deep image deblurring'' by \cite{SRN}
  \vspace{0.5mm}
  \\
  \hline
  DCP & ``Single image haze removal using dark channel prior'' by \cite{dehaze6}
  \vspace{0.5mm}
  \\
  NLD & ``Non-local image dehazing'' by \cite{NLD}
  \vspace{0.5mm}
  \\
  AODN & ``Aod-net: All-in-one dehazing network'' by \cite{AODnet}
  \vspace{0.5mm}
  \\
  DGFN & MSE-based DGFN in ``Gated fusion network for single image dehazing'' by \cite{GFN}
  \vspace{0.5mm}
  \\
  GCANet & ``Gated Context Aggregation Network for Image Dehazing and Deraining'' by \cite{GCANet}
  \\
  PFFNet & ``Progressive feature fusion network for realistic image dehazing'' by \cite{dehaze12}
  \vspace{0.5mm}
  \\
  \hline
  RESN & ``Recurrent squeeze-and-excitation context aggregation net for single image deraining'' by \cite{RESCAN}
  \vspace{0.5mm}
  \\
  IDGAN & ``Image de-raining using a conditional generative adversarial network'' by \cite{ID-CGAN}
  \vspace{0.5mm}
  \\
  DID-MDN & ``Density-aware single image de-raining using a multi-stream dense network'' by \cite{DID-MDN}
  \vspace{0.5mm}
  \\
  \hline
\end{tabular}
\end{adjustbox}
  \label{tab:list}
  \vspace{-2mm}
  \end{table*}

\subsection{\textbf{Additional Visual Results}} 
\label{subsec:supp-C}
\begin{figure*}[tb]
  \small
  \centering
  \begin{tabular}{cccc}
    \hspace{-3mm}
     \includegraphics[width=0.245\linewidth]{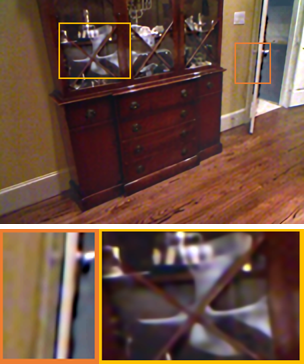} & \hspace{-4mm}
      \includegraphics[width=0.245\linewidth]{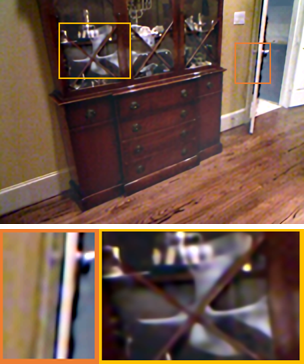} & \hspace{-4mm}
      \includegraphics[width=0.245\linewidth]{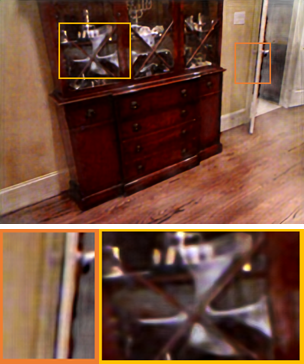} & \hspace{-4mm}
      \includegraphics[width=0.245\linewidth]{00032_GFN_x4.png}
    \vspace{-0.5mm}
    \\
    \hspace{-3mm} 
    (a) EDSR + DCP & \hspace{-4mm}
    (b) EDSR + NLD & \hspace{-4mm}
    (c) EDSR + DGFN & \hspace{-4mm}
    (d) GFN (ours)
    \\
    \hspace{-3mm} 
    18.68 / 0.823 & \hspace{-4mm}
    16.04 / 0.761 & \hspace{-4mm}
    17.54 / 0.875 & \hspace{-4mm}
    28.85 / 0.961
  \end{tabular}
  \caption{
    \textbf{More visual comparisons on the LR-RESIDE dataset.}
  The evaluated methods include EDSR \citep{EDSR}, DCP \citep{dehaze6}, NLD \citep{NLD}, and DGFN \citep{GFN}.
  The proposed model generates sharper HR images with more details.
  }
  \label{fig:visual_results_RESIDE_supp}
  \end{figure*}

In this section, we present more qualitative comparisons on the LR-RESIDE in \figref{visual_results_RESIDE_supp}, which includes the combinations of the SR algorithm \citep{EDSR} and 
dehazing algorithms \citep{dehaze6, NLD, GFN}.

  \subsection{\textbf{Additional Ablation Study and analysis}} 
  \label{subsec:supp-D}
  To further demonstrate the importance of the dual-branch architecture and gate module, more ablation study and visual results are presented in this section.
  We first compare the restoration module with the state-of-the-art restoration methods to evaluate the performance contribution brought by the image restoration module.
  Then, the qualitative results of the ablation study are presented to demonstrate how other modules help to improve the performance.

  {\flushleft \bf Performance of Restoration Module.}
  We provide the quantitative results from the state-of-the-art restoration methods and the proposed restoration module in \tabref{restoration-module-test}.
  The restoration methods include deblurring algorithms (DeepDeblur \cite{deepdeblur}, DeblurGAN \cite{deblurgan}, and SRN \cite{SRN}),
  dehazing  algorithms (DGFN \cite{GFN}, GCANet \cite{GCANet}, and PFFNet \cite{dehaze12}), 
  and deraining algorithms (IDGAN \cite{ID-CGAN}, RESN \cite{RESCAN}, and DID-MDN \cite{DID-MDN}).
  Since these restoration methods are trained on the high-resolution images (GOPRO, RESIDE, \\
  Rain1200 datasets), 
  we re-train the restoration module on the same high-resolution datasets for fair comparisons.
  As shown in \tabref{restoration-module-test}, 
  in none of these three datasets does our restoration module acquire the best results, 
  while the proposed GFN still performs favorably on all the three datasets as shown in Table 1,2,3 of the manuscript.
  Therefore, the favorable performance of the proposed method comes from the architecture designs, such as the dual-branch architecture and the gate module. 

\begin{figure*}[tb]
    \small
    \centering
    \begin{tabular}{ccccc}
      \hspace{-3mm}
        \includegraphics[width=0.19\linewidth]{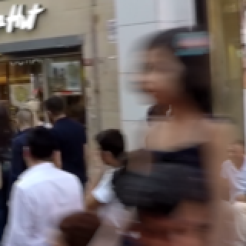} & \hspace{-4mm}
        \includegraphics[width=0.19\linewidth]{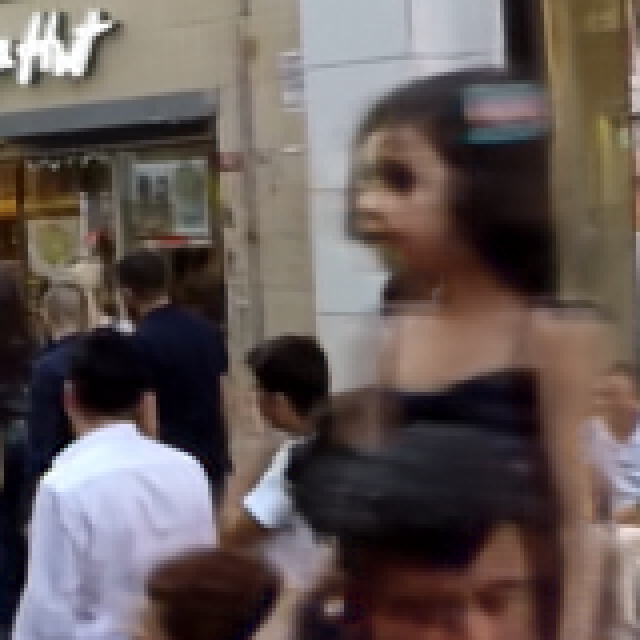} & \hspace{-4mm}
        \includegraphics[width=0.19\linewidth]{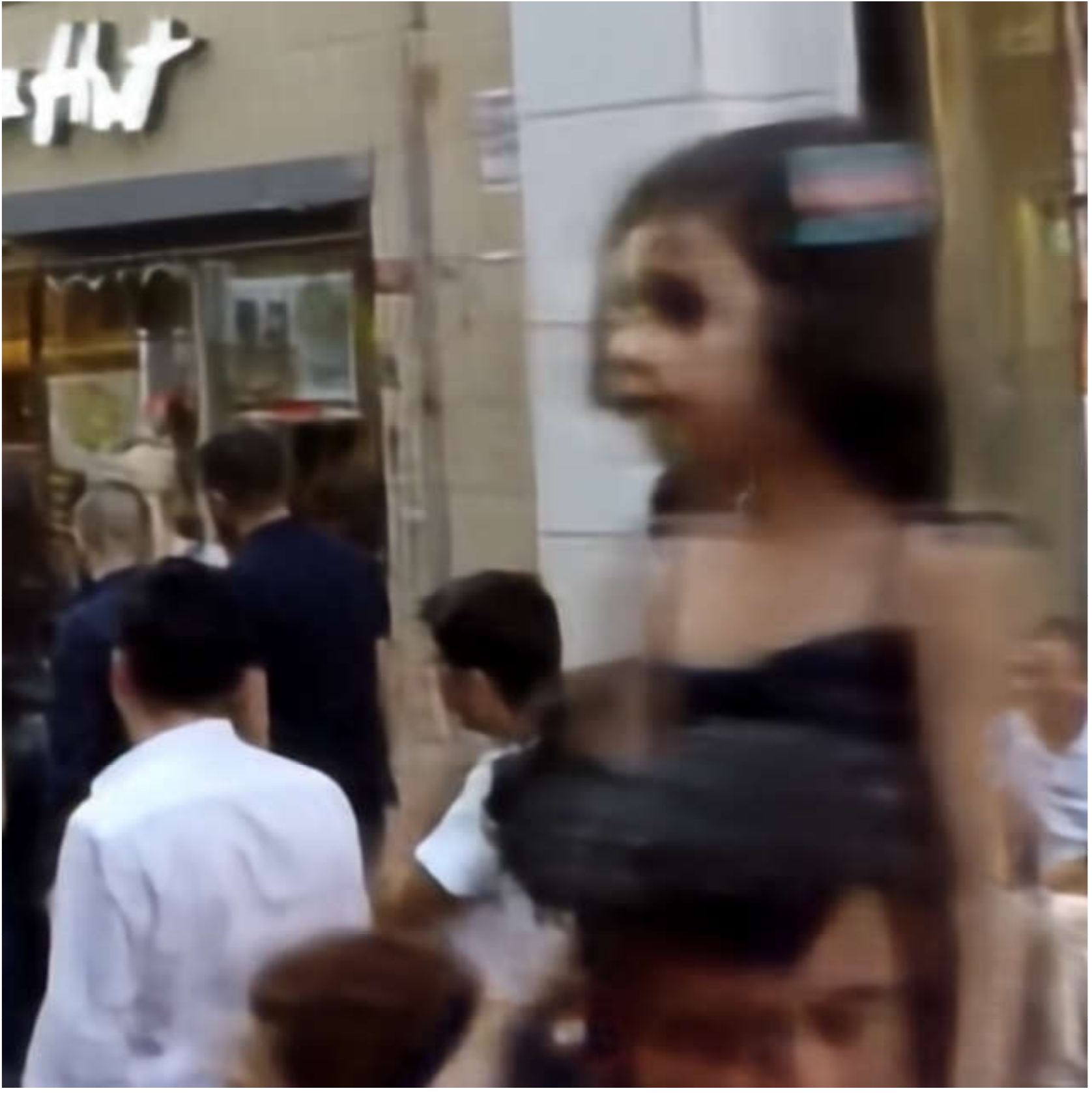} & \hspace{-4mm}
        \includegraphics[width=0.19\linewidth]{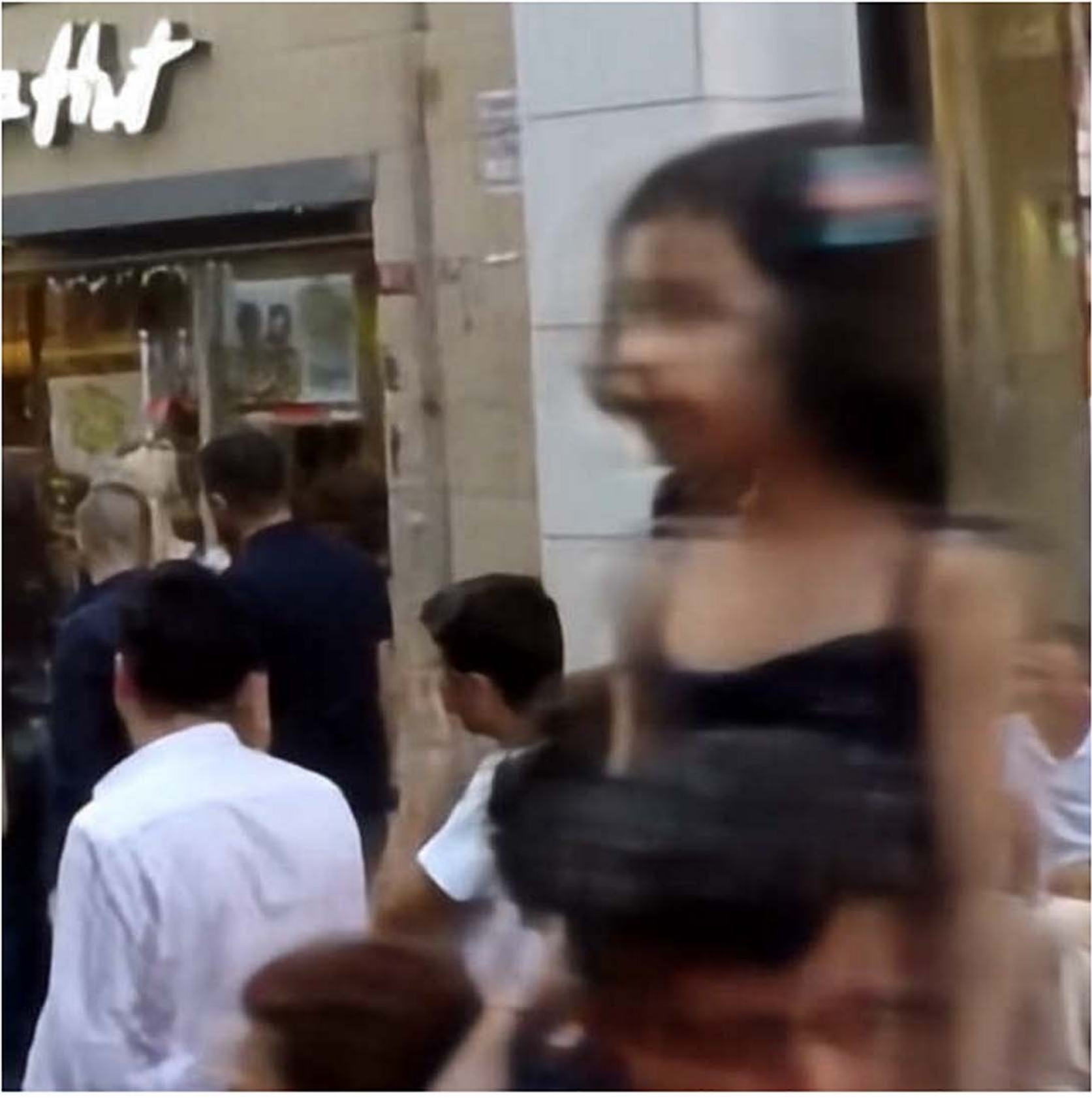} & \hspace{-4mm}
        \includegraphics[width=0.19\linewidth]{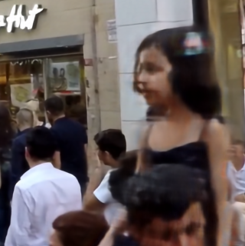}
      \vspace{-0.5mm}
      \\
      \hspace{-3mm} 
        (a) Blurry LR input & \hspace{-4mm}
        (b) $G_{res}$ in Model-1 & \hspace{-4mm}
        (c) Model-1 & \hspace{-4mm}
        (d) GFN w/o $G_{gate}$ & \hspace{-4mm}
        (e) GFN
      \\
      \hspace{-3mm}
        \includegraphics[width=0.19\linewidth]{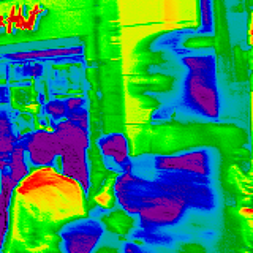} & \hspace{-4mm}
        \includegraphics[width=0.19\linewidth]{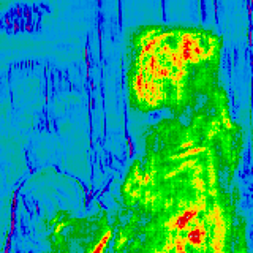} & \hspace{-4mm}
        \includegraphics[width=0.19\linewidth]{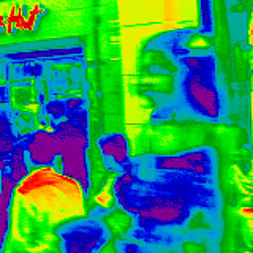} & \hspace{-4mm}
        \includegraphics[width=0.19\linewidth]{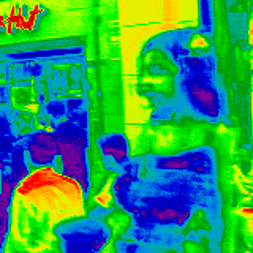} & \hspace{-4mm}
        \includegraphics[width=0.19\linewidth]{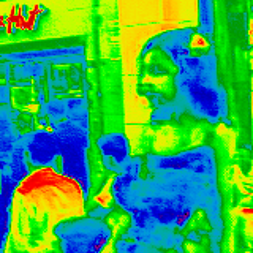}
        \vspace{-0.5mm}
      \\
      \hspace{-3mm} 
        (f) $\phi_{BF}$ & \hspace{-4mm}
        (g) $\phi_{RF}$ & \hspace{-4mm}
        (h) $\phi{}^{1}_{fusion}$ & \hspace{-4mm}
        (i) $\phi{}^{2}_{fusion}$ & \hspace{-4mm}
        (j) $\phi{}^{3}_{fusion}$
    \end{tabular}
    \caption{
      \textbf{Qualitative results of the ablation study.}
      $\phi_{BF}$ denotes the base features from the base module $G_{base}$ and $\phi_{RF}$ denotes the features from the restoration module $G_{res}$.
      All the models are trained on the LR-GOPRO dataset with the same training settings as the proposed GFN.
    }
    \label{fig:artifacts_suppression}
    \end{figure*}

  {\flushleft \bf Effect of dual-branch architecture and gate module.}
  To further demonstrate the benefits of the dual-branch architecture and gate module, we present an example in \figref{artifacts_suppression}.
  \figref{artifacts_suppression}(b) and (c) show 
  the outputs of the restoration module $G_{res}$ and Model-1 ($G_{res}$ + $G_{base}$ + $G_{recon}$) in \figref{supp_ablation}(a).
  Since the artifacts in the $G_{res}$ are propagated to the $G_{base}$ and $G_{recon}$, the Model-1 generates less satisfactory results as shown in \figref{artifacts_suppression}(c).
  \figref{artifacts_suppression}(d) shows the output of the Model-4 in \figref{supp_ablation}(d), which adopts the dual-branch architecture without the gate module $G_{gate}$.
 \figref{artifacts_suppression}(d) contains fewer artifacts than \figref{artifacts_suppression}(c), especially on the regions that are relatively sharper in the input image. 
  This is because the dual-branch architecture combines features from both input images and recovered images and, therefore, avoids error propagation from only the recovered images.
  \figref{artifacts_suppression}(e) shows the output of the proposed GFN introducing the gate module to adaptively fuse the features.
  By exploiting the confidence of the features from two branches ($\phi_{RF}$ into $\phi_{BF}$), the gate module manages to suppress the artifacts and blurry features via local and channel-wise feature fusion.
\figref{artifacts_suppression}(f)-(j) shows that our model progressively fuses features and suppresses artifacts through the gate module.

  \begin{table*}[tb]
    \small
    \centering
    \caption{
      \textbf{Quantitative comparison with the state-of-the-art restoration methods on three applications.}
      All the comparison methods for each application are trained using the same setting.
      {\color{red}\textbf{Red texts}} and {\color{blue}blue texts} indicate the best and the second-best performance respectively.
    }
    \begin{tabular}[t]{ccccc}
    \multicolumn{5}{c}{\textbf{GOPRO dataset}} 
    \\
    \hline
    \multirow{-1}{*}{}  &Restoration Module  &DeepDeblur &DeblurGAN &~~~SRN~~~ 
    \\ 
    \hline
    Deblurring PSNR                     &{\color{blue}29.16}	&27.48	&27.02	 &{\color{red}\textbf{30.26}}		 
    \vspace{0.5mm}
    \\
    \hline
    \\
    \multicolumn{5}{c}{\textbf{RESIDE dataset}}
    \\
    \hline
    \multirow{-1}{*}{}  &Restoration Module  &DGFN &GCANet &PFFNet 
    \\ 
    \hline
    Dehazing PSNR                     &24.46	&23.47	&{\color{blue}26.32}	 &{\color{red}\textbf{28.20}}		  
    \vspace{0.5mm}
    \\
    \hline
    \\
    \multicolumn{5}{c}{\textbf{Rain1200 dataset}}
    \\
    \hline
    \multirow{-1}{*}{}  &Restoration Module  &IDGAN &RESN &DID-MDN 
    \\ 
    \hline
    Deraining PSNR                     &{\color{blue}29.36}	&27.50	&29.12	 &{\color{red}\textbf{30.10}}		 
    \vspace{0.5mm}
    \\
    \hline

    \end{tabular}
    \label{tab:restoration-module-test}
    \vspace{-2mm}
    \end{table*}

  \subsection{\textbf{Applications on Detection Task}} 
  \label{subsec:supp-E}
  To demonstrate that the proposed method can help the following high-level tasks, we compare the proposed GFN with state-of-the-art methods on the object detection task.
We first generate two datasets from the KITTI dataset \citep{kitti_detection}, one blurry low-resolution dataset and a hazy low-resolution dataset.
 For the blurry dataset, we apply the single image non-uniform blurry synthesis method in \cite{blur_sythn} to generate the blurry HR images and use the bicubic downsampling to generate the blurry LR images as the inputs.
  We then generate recovered HR images with the following methods: the bicubic upsampling, deblurring method SRN \citep{SRN} with super resolution method RCAN \citep{RCAN}, joint restoration and super-resolution method ED-DSRN \citep{icassp18}, 
  and the proposed GFN.
  For the hazy dataset, we first apply the single image depth estimation method, the Monodepth2 \citep{monodepth2}, to predict a depth map for each image and then synthesize the hazy image following the instruction of the RESIDE dataset \cite{RESIDE}.
  We compare the proposed GFN with the following approaches: the bicubic upsampling, dehazing method PFFNet \citep{dehaze12} with super resolution method RCAN \citep{RCAN}, and joint restoration and super-resolution method ED-DSRN \citep{icassp18}.
  We use the above methods to recover HR images and then use the YOLOv3 \citep{yolov3} to evaluate the detection accuracy.  %

\begin{figure*}[tb]
    \small
    \centering
    \begin{tabular}{ccc}
      \hspace{-3mm}
        \includegraphics[width=0.32\linewidth]{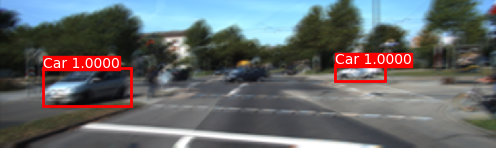} & \hspace{-4mm}
        \includegraphics[width=0.32\linewidth]{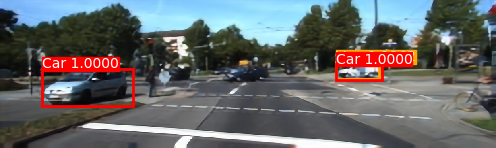} & \hspace{-4mm}
        \includegraphics[width=0.32\linewidth]{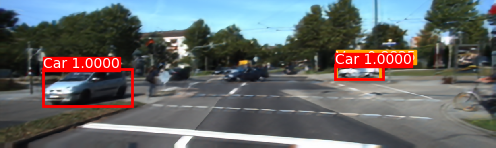} 
      \vspace{-0.5mm}
      \\
      \hspace{-3mm} 
        (a) Bicubic & \hspace{-4mm}
        (b) SRN+RCAN & \hspace{-4mm}
        (c) RCAN+SRN

      \\
      \hspace{-3mm}
        \includegraphics[width=0.32\linewidth]{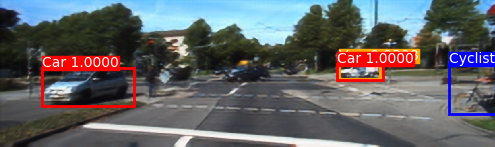} & \hspace{-4mm}
        \includegraphics[width=0.32\linewidth]{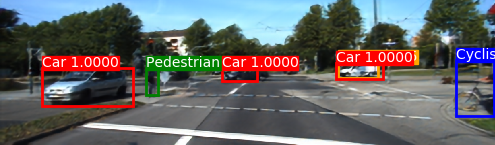} & \hspace{-4mm}
        \includegraphics[width=0.32\linewidth]{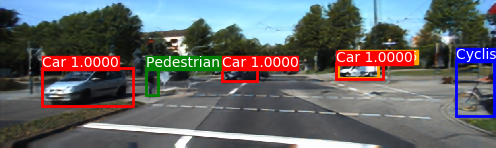} 
        \vspace{-0.5mm}
      \\
      \hspace{-3mm} 
        (d) ED-DSRN & \hspace{-4mm}
        (e) GFN~(ours) & \hspace{-4mm}
        (f) Ground-truth HR

    \end{tabular}
    \caption{
    \textbf{Detection results using the recovered images from different methods. }
    We compare the following methods: bicubic upsampling, deblurring method SRN \citep{SRN} + super resolution method RCAN \citep{RCAN}, joint restoration and super-resolution method ED-DSRN \citep{icassp18}, and the proposed GFN.
    }
    \label{fig:visual_results_detection_db}
    \end{figure*}

\begin{figure*}[tb]
      \small
      \centering
      \begin{tabular}{ccc}
        \hspace{-3mm}
          \includegraphics[width=0.32\linewidth]{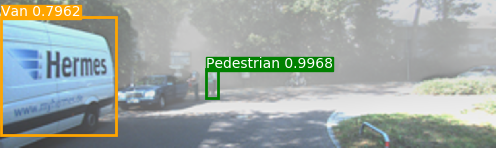} & \hspace{-4mm}
          \includegraphics[width=0.32\linewidth]{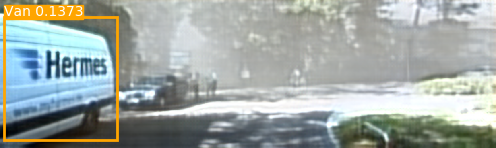} & \hspace{-4mm}
          \includegraphics[width=0.32\linewidth]{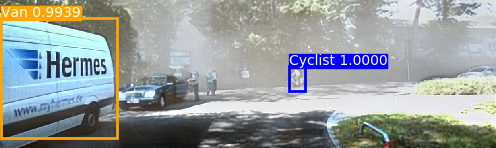} 
        \vspace{-0.5mm}
        \\
        \hspace{-3mm} 
          (a) Bicubic & \hspace{-4mm}
          (b) PFFNet+RCAN & \hspace{-4mm}
          (c) RCAN+PFFNet

        \\
        \hspace{-3mm}
          \includegraphics[width=0.32\linewidth]{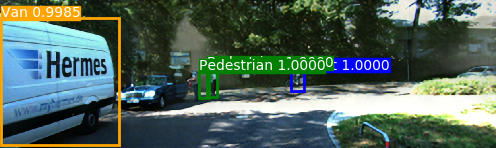} & \hspace{-4mm}
          \includegraphics[width=0.32\linewidth]{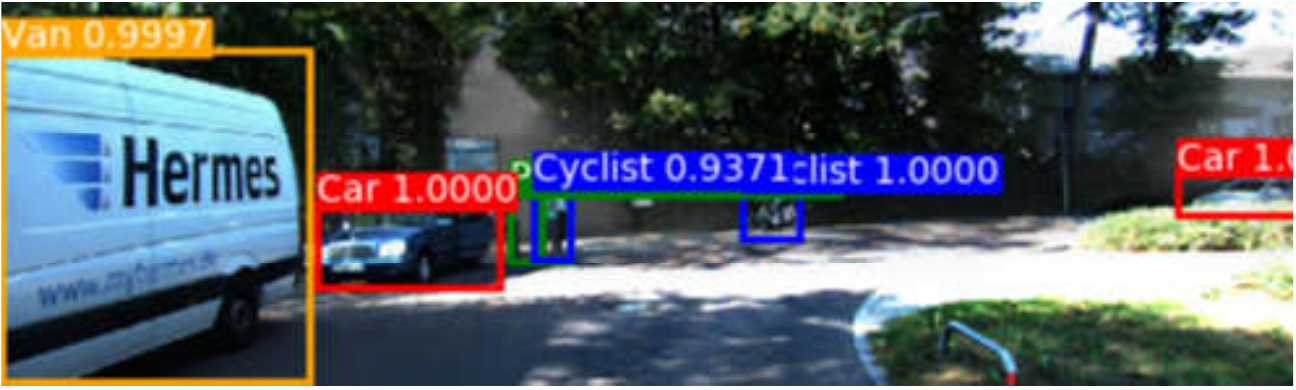} & \hspace{-4mm}
          \includegraphics[width=0.32\linewidth]{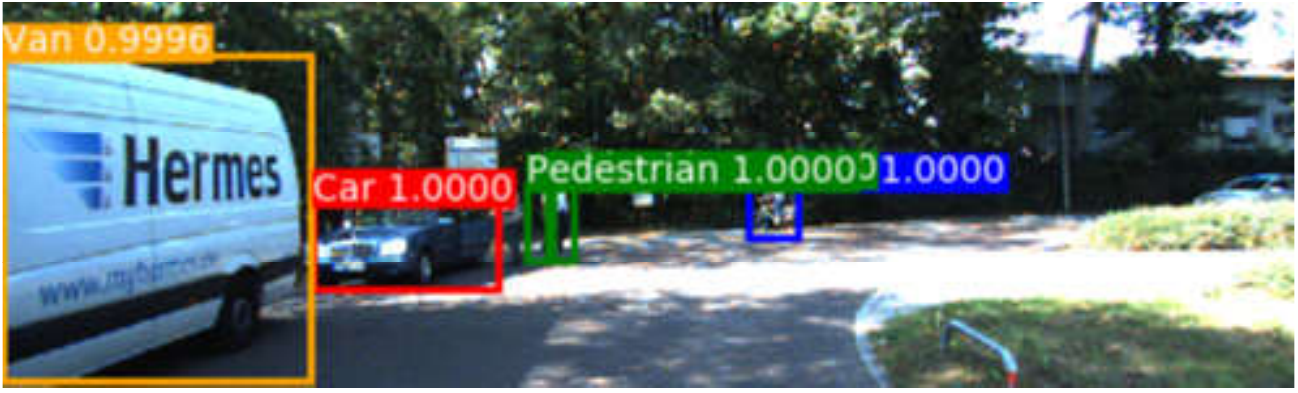} 
          \vspace{-0.5mm}
        \\
        \hspace{-3mm} 
          (d) ED-DSRN & \hspace{-4mm}
          (e) GFN~(ours) & \hspace{-4mm}
          (f) Ground-truth HR

      \end{tabular}
      \caption{
      \textbf{Detection results using the recovered images from different methods.}
      We compare the following methods: bicubic upsampling, dehazing method PFFNet \citep{dehaze12} + super resolution method RCAN \citep{RCAN}, joint restoration and super-resolution method ED-DSRN \citep{icassp18}, and the proposed GFN.
      }
      \label{fig:visual_results_detection_dh}
      \end{figure*}

We show the detection accuracy in \tabref{high-level-detection-blur} and \tabref{high-level-detection-hazy}.
 The HR images restored from the proposed GFN obtain the best detection accuracy in both applications.
 The qualitative results in \figref{visual_results_detection_db} and \figref{visual_results_detection_dh} demonstrate that 
 our GFN not only generates clean HR outputs but also improves the detection algorithm to recognize the cars and pedestrians. 

 \begin{table*}[]
  \small
  \centering
  \caption{
  \textbf{Objects detection results on the KITTI detection dataset \citep{kitti_detection} with non-uniform motion blur.}
  We test different joint deblurring and super-resolution methods, and use YOLOv3 \citep{yolov3} as the detection algorithm.
  The comparison methods include bicubic upsampling, deblurring method SRN \citep{SRN} + super resolution method RCAN \citep{RCAN}, joint restoration and super-resolution method ED-DSRN \citep{icassp18}, and the proposed GFN.
  We also show the detection result using the ground-truth sharp HR image.
  The mAP is the abbreviation of mean average precision.
  {\color{red}\textbf{Red texts}} indicate the best detection precision except for the Ground-truth HR.
  }
  \begin{tabular}[t]{c|cccccc}
  \hline
  \multirow{-1}{*}{YOLOv3}  &Bicubic  &SRN+RCAN &RCAN+SRN &~~~ED-DSRN~~~ &~~~GFN~~~ &Ground-truth HR
  \\ 
  \hline
  Car                          &0.258	&0.481  &0.481	&0.416	 &0.499		&0.812  
  \\
  Van                          &0.149	&0.358  &0.392	&0.298	 &0.406		&0.724 
  \\
  Truck                        &0.208	&0.558  &0.578	&0.499	 &0.584		&0.842 
  \\
  Pedestrian                   &0.164	&0.327  &0.329	&0.305	 &0.370		&0.604 
  \\
  Person Sitting               &0.028 &0.187  &0.105	&0.122	 &0.163		&0.436 
  \\
  Cyclist                      &0.026	&0.203  &0.171	&0.158	 &0.283		&0.592 
  \\
  Tram                         &0.108 &0.331  &0.314	&0.272	 &0.383		&0.796 
  \\
  \hline
  \textbf{mAP}                 &0.120	&0.316  &0.308	&0.267	 &{\color{red}\textbf{0.352}}		&0.646
  \\
  \hline
  \end{tabular}
  \label{tab:high-level-detection-blur}
  \vspace{-2mm}
  \end{table*}

\begin{table*}[]
  \small
  \centering
  \caption{
  \textbf{Objects detection results on the KITTI detection dataset \citep{kitti_detection} with haze degradation.}
  We test different joint dehazing and super-resolution methods, and use YOLOv3 \citep{yolov3} as the detection algorithm.
  The comparison methods include bicubic upsampling, dehazing method PFFNet \citep{dehaze12} + super resolution method RCAN \citep{RCAN}, joint restoration and super-resolution method ED-DSRN \citep{icassp18}, and the proposed GFN.
  We also show the detection result using the ground-truth sharp HR image.
  The mAP is the abbreviation of mean average precision.
  {\color{red}\textbf{Red texts}} indicate the best detection precision except for the Ground-truth HR.
  }
  \begin{tabular}[]{c|cccccc}
  \hline
  \multirow{-1}{*}{YOLOv3}  &Bicubic  &PFFNet+RCAN &RCAN+PFFNet &~~~ED-DSRN~~~ &~~~GFN~~~ &Ground-truth HR
  \\ 
  \hline
  Car                          &0.146	&0.073  &0.254	&0.431	 &0.505		&0.812  
  \\
  Van                          &0.053	&0.033  &0.113	&0.228	 &0.301		&0.724 
  \\
  Truck                        &0.033	&0.008  &0.047	&0.119	 &0.197		&0.842 
  \\
  Pedestrian                   &0.178	&0.067  &0.297	&0.406	 &0.461		&0.604 
  \\
  Person Sitting               &0.000 &0.000  &0.021	&0.204	 &0.327		&0.436 
  \\
  Cyclist                      &0.055	&0.027  &0.179	&0.196	 &0.279		&0.592 
  \\
  Tram                         &0.000 &0.000  &0.007	&0.097	 &0.239		&0.796 
  \\
  \hline
  \textbf{mAP}                 &0.058	&0.026  &0.117	&0.219	 &{\color{red}\textbf{0.303}}		&0.646
  \\
  \hline
  \end{tabular}
  \label{tab:high-level-detection-hazy}
  \vspace{-2mm}
  \end{table*}

\end{document}